%% file: main.tex
\documentclass{article}
\usepackage[numbers]{natbib}  

\usepackage[main, final]{neurips_2026}

\usepackage[utf8]{inputenc} 
\usepackage[T1]{fontenc}    
\usepackage{hyperref}       
\usepackage{url}            
\usepackage{booktabs}       
\usepackage{amsfonts}       
\usepackage{nicefrac}       
\usepackage{microtype}      
\usepackage{xcolor}         

\usepackage{graphicx} 
\usepackage[inline]{enumitem}

\usepackage{algorithm}
\usepackage{algorithmic}

\usepackage{amsmath}
\usepackage{amsfonts}
\usepackage{amssymb}
\usepackage{amsthm}
\usepackage{multirow}     
\usepackage{colortbl}     

\usepackage{wrapfig}
\usepackage{caption} 

\usepackage{minitoc}

\newtheorem{theorem}{Theorem}
\newtheorem{proposition}{Proposition}
\newtheorem{definition}{Definition}
\newtheorem{corollary}{Corollary}

\DeclareMathOperator*{\argmin}{arg\,min}

\input{macros.tex}

\title{A Sparse Low-Rank Biclique Decomposition for Graphs}

\author{%
Antoine Vialle\thanks{Corresponding authors: Antoine Vialle and Jhony H. Giraldo.} \\
LTCI, Télécom Paris, IP Paris \\
Palaiseau, France \\
\texttt{antoine.vialle@telecom-paris.fr}
\And
Sergei Gerasimov \\
LTCI, Télécom Paris, IP Paris \\
Palaiseau, France \\
\texttt{sergei.gerasimov@telecom-paris.fr}
\And
Aref Einizade \\
SAMOVAR, Télécom SudParis, IP Paris \\
Palaiseau, France \\
\texttt{aref.einizade@telecom-sudparis.eu}
\And
Antonio Ortega \\
University of Southern California \\
Los Angeles, USA \\
\texttt{aortega@usc.edu}
\AND
Fragkiskos D. Malliaros \\
Université Paris Saclay, CentraleSupélec, Inria \\
Gif-Sur-Yvette, France \\
\texttt{fragkiskos.malliaros@centralesupelec.fr}
\And
Jhony H. Giraldo\footnotemark[1] \\
LTCI, Télécom Paris, IP Paris \\
Palaiseau, France \\
\texttt{jhony.giraldo@telecom-paris.fr}
}

\begin{document}

\maketitle

\begin{abstract}

Low-rank plus sparse decompositions, such as robust PCA, typically model the low-rank component as dense and the sparse component as corruption. This viewpoint is poorly suited to graph adjacency matrices, where sparsity is structural and dense low-rank factors destroy the combinatorial and computational properties of the graph. We propose signed biclique decomposition (SBD), a sparse low-rank decomposition for integer-weighted graphs that represents the structured component as a sum of sparse signed binary rank-one factors, together with a sparse signed residual. The resulting representation remains discrete, interpretable, sparse, and compatible with standard linear algebra. We formulate fixed-rank and greedy SBD objectives and derive a gain-based greedy algorithm with theoretical guarantees. Empirically, SBD recovers block structure in noisy graphs, while its factorized representation enables efficient sparse matrix-matrix multiplication (SpMM). On real graphs with strong shared-neighborhood structure, SBD achieves compression factors above $10$ and SpMM speedups up to $6\times$ over cuSPARSE, suggesting a practical alternative to dense low-rank decompositions for graphs. Code is available at \href{https://github.com/IAntoineV/sbd}{github.com/IAntoineV/sbd}.

\end{abstract}

\section{Introduction}

Low-rank decomposition is a fundamental tool for uncovering latent structure in high-dimensional data. 
In sparse plus low-rank (SLR) models \citep{Recht2010} {(see Table~\ref{tab:abbreviations} for abbreviations)}, a data matrix $\Am$ is decomposed as $\Am = \Lm_0 + \Rm_0$, where $\Lm_0$ is a low-rank matrix that captures global or latent structure, and $\Rm_0$ is a sparse matrix capturing localized deviations, outliers, or corruptions. We refer to $\Rm_0$ as the \emph{residual} (or remainder) of the decomposition.
A prominent example of SLR is robust principal component analysis (RPCA) \citep{RPCA}, which recovers these two components through a convex optimisation problem. RPCA and related SLR methods have been successful in many applications, including medical imaging \citep{mrm_RPCA, Chen2017LowRankSparseMRI}, asset pricing \citep{Basu2019}, background subtraction, and corruption removal \citep{RPCA}.

In this paper, we ask what an SLR-like decomposition should look like for sparse graph adjacency matrices. Since sparsity is intrinsic to graph structure, both components should remain sparse: the structured term should capture shared connectivity patterns through graph-supported objects, such as clique-, biclique-, or shared-neighborhood blocks, while the residual should store sparse signed corrections. Classical SLR methods such as RPCA are poorly matched to this goal: their low-rank component is typically dense and incoherent, and sparsity is treated as corruption \citep{RPCA}. Applied to graphs, this can introduce dense fill-in, destroy the adjacency sparsity pattern, and produce representations that are inefficient to store (see Appendix \ref{app:slr}).

A relevant alternative is sparse PCA (SPCA) \citep{SPCA2006, SCoTLASS2003, parekh2017improved}, which enforces sparsity on the low-rank factors rather than directly on the reconstructed matrix or residual.
However, the resulting singular vectors are generally not binary or indicator-like, and thus the induced low-rank approximation is poorly aligned with the discrete structure of graphs. As a consequence, the residual component typically remains comparably dense, offering limited improvement in sparsity over the original adjacency matrix $\Am$ (see Appendix \ref{app:slr}).

\begin{figure}
    \centering
    \includegraphics[width=\linewidth]{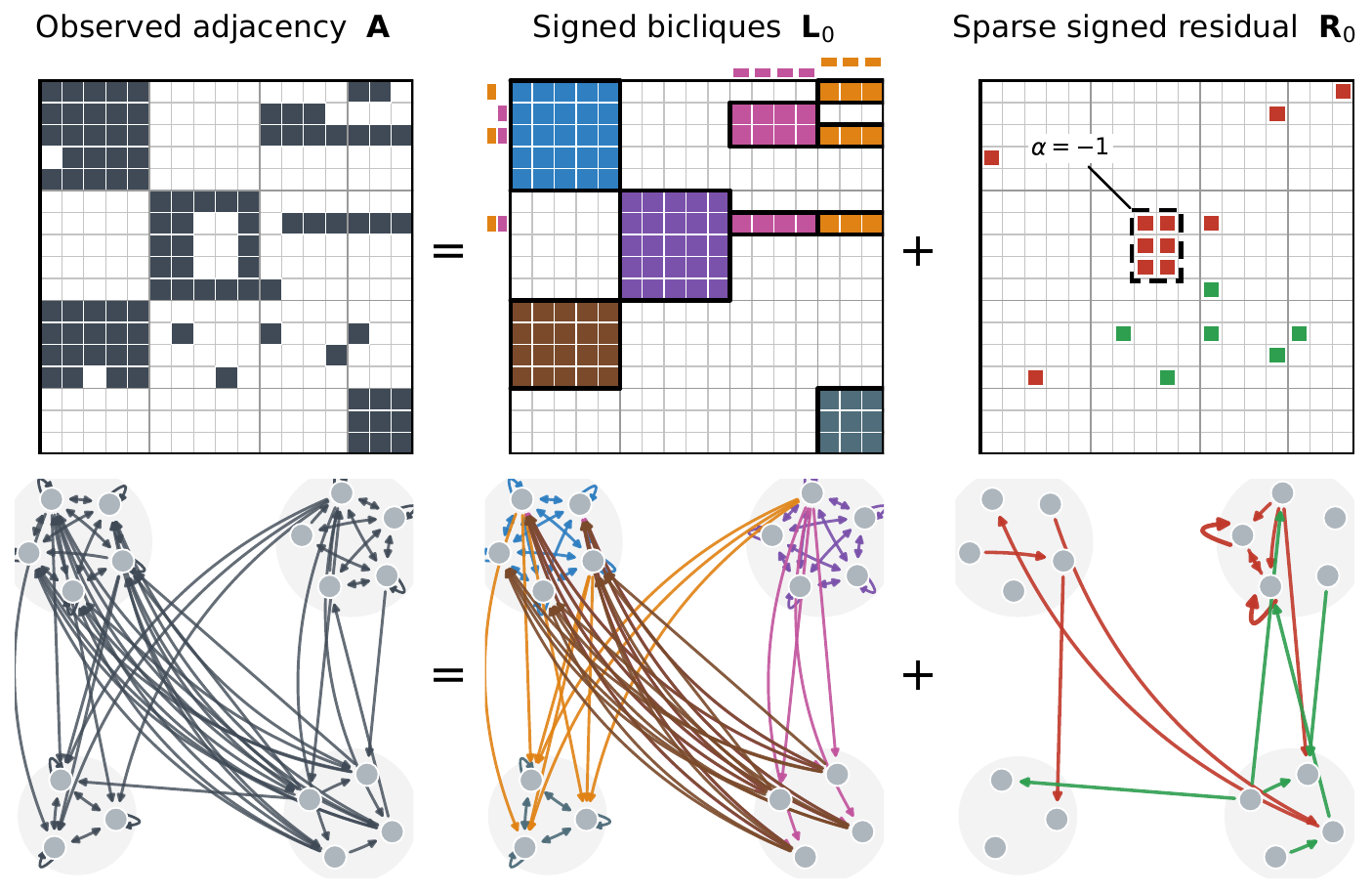}
    \caption{A sparse graph adjacency matrix $\Am$ is decomposed into
    a sparse low-rank block component $\mathbf L_0$ and a sparse signed residual $\Rm_0$. Green entries denote added edges, while red entries denote missing edges relative to the recovered block structure.}
    \label{fig:teaser}
\end{figure}

These limitations suggest a different sparse-low-rank principle for graphs: the structured component should be both low-rank and sparse, while the residual should remain sparse. 
This leads to a sparse plus low-rank-and-sparse (SLRS) formulation, in which $\Lm_0$ is represented by a small number of sparse graph-supported low-rank components and $\Rm_0$ stores the remaining signed corrections.
Figure \ref{fig:teaser} illustrates this goal: a sparse graph adjacency matrix is decomposed into a sparse low-rank block component and a sparse signed residual. This is the formulation pursued in this work.

In this paper, we propose sparse low-rank decompositions for integer-weighted graphs from a combinatorial perspective. 
We establish a connection between RPCA-style decompositions and sparse decompositions of graphs by introducing a signed biclique decomposition (SBD) framework.
SBD represents a graph using a residual and a small number of sparse, signed, binary components, which can be interpreted as discrete shared-neighborhood or biclique-like patterns.
This result provides a theoretical foundation for fast and scalable SLRS models on graphs and shows that low-rank graph structure can be approximated by discrete combinatorial objects rather than dense real-valued factors.
Building on this connection, we propose a simple greedy algorithm that extracts such components efficiently and scales to large graphs.

SBD is motivated by the need for graph representations that are simultaneously sparse, interpretable, and computationally efficient. Sparser representations are desirable in graph settings as they preserve locality and reduce redundancy in the underlying structure. Empirically, we show that this formulation matches the reconstruction behavior of RPCA-style methods on small graphs while remaining scalable to large-scale settings. 
Moreover, the resulting decomposition enables significant computational gains, in particular, by accelerating sparse matrix-matrix multiplication (SpMM).

Our main contributions are:
\begin{itemize}[leftmargin=0.5cm]
    \itemsep0em
    \item We formulate graph sparse low-rank decomposition as a signed biclique decomposition of an integer-weighted adjacency matrix. Each component is a sparse signed binary rank-one matrix, so the resulting low-rank term remains combinatorial and sparse rather than dense.
    \item We derive a greedy objective for signed bicliques. We prove that positive-gain steps decrease the residual norm and give conditions for alternating optimisation to find them. We bound the remainder using the stopping rule.
    \item We introduce scalable SBD implementations for medium and large graphs. Because the decomposition is expressed in standard linear algebra, it can be used directly to accelerate SpMM.
    \item We evaluate the method on graph denoising and\ SpMM benchmarks, where the sparse signed-biclique representation yields speedups over standard sparse kernels on graphs with strong shared-neighborhood structure.
\end{itemize}

\section{Related Work}

Several lines of work are closely related to our sparse low-rank decomposition framework.

\textbf{Sparse low-rank decompositions for graphs and matrices.}
The most closely related works study matrices that are simultaneously sparse and low-rank. 
\citet{Savalle2012EstimationOS} consider the estimation of matrices that satisfy both constraints, motivating sparse low-rank structure beyond the classical RPCA setting. 
Closer to our setting, \citet{Kanada2018} propose a low-rank sparse decomposition of graph adjacency matrices for extracting clean clusters. 
These works share our motivation that graph structure should not be represented by a dense low-rank matrix. Like RPCA, these methods use singular-value thresholding for regularized nuclear-norm minimization, which is difficult to scale to large graphs and does not provide the discrete signed-biclique representation considered here.

\textbf{Binary and Boolean matrix decompositions.}
A related line of work decomposes binary matrices under algebraic structures, such as the Boolean semiring or arithmetic over $\mathbb{Z}/2\mathbb{Z}$ \citep{Miettinen2008, Dan2018, BELOHLAVEK20103, BMF2024}. 
These models suit compression and pattern discovery, but they use notions of rank that differ from standard linear algebra: Boolean addition satisfies $1+1=1$, while arithmetic over $\mathbb{Z}/2\mathbb{Z}$ satisfies $1+1=0$. 
Consequently, they do not directly support linear operations such as multiplying an adjacency matrix by a dense feature matrix. Binary matrix factorization methods use ordinary arithmetic \citep{proximus2003,Zhang2007, Shen2009, BECKERLEG2020}, this typically restricts the model to unsigned bicliques, making it closer to edge partitioning than to a signed low-rank-plus-residual decomposition. Other methods, like \citet{Miller2009}, learn binary latent features on nodes to predict edge probabilities of the graph.

\textbf{Cut, biclique, and spectral graph decompositions.}
Cut and biclique decompositions have also been studied in graph approximation and regularity theory. 
The Frieze--Kannan weak regularity lemma \citep{FriezeKannan1, FriezeKannan2} approximates a weighted adjacency matrix by a sum of weighted cut matrices, equivalently biclique-like blocks, with a residual controlled in cut norm. 
This provides an important connection between matrix approximation and graph structure. 
However, regularity-based decompositions are not designed to preserve sparsity.

There is also a classical connection between biclique decompositions and spectral graph structure. 
For example, the minimum number of bicliques required to partition the edges of a graph is lower bounded by $\max(n_+(\Am), n_-(\Am))$, where $n_+(\Am)$ and $n_-(\Am)$ denote the numbers of positive and negative eigenvalues of the adjacency matrix \citep{GREGORY1999, Graham1971}. 
This shows that the biclique structure is closely tied to the spectrum of the graph. 
Related work by \citet{Ames2011} studies recovery of planted cliques and bicliques via nuclear norm minimization, further connecting combinatorial graph patterns with low-rank structure. 
These results motivate the use of biclique-like components.

\textbf{Graph compression and summarization.}
Graph compression and summarization methods aim to represent graphs using compact structures, often through lossless or near-lossless encodings \citep{Liu2018, Maciej2018}. While these methods are effective for reducing storage or identifying dense patterns, they are not designed to provide low-rank plus sparse decompositions compatible with linear algebraic operations. Those are much more aligned with binary matrix decomposition.

\textbf{Accelerating SpMM on graphs.} Several optimized GPU kernels have been proposed to accelerate SpMM on graphs, including Sputnik~\citep{sgk_sc2020}, dgSparse, and GE-SpMM~\citep{9355302}. More recently, compressed binary matrices (CBM)~\citep{Joao2025} proposed compressing graph adjacency operators using ordered neighborhood differences and cumulative sums to accelerate SpMM.

Overall, prior work either provides sparse low-rank formulations without a scalable discrete signed representation, binary decompositions under nonstandard algebraic operations, or graph compression methods that are not designed as linear low-rank-plus-residual models. 
Our contribution is to combine these desiderata: we decompose an integer-valued adjacency matrix in standard linear space into sparse signed biclique components and a sparse signed residual. 
The signed components allow cancellations between overlapping structures, while the binary supports preserve a combinatorial graph interpretation and enable efficient storage and computation.

\section{Preliminaries}

\textbf{Notation.}
We write vectors and matrices in boldface. For an integer $n \geq 1$, let
$[n] := \{1,\ldots,n\}$. For a matrix
$\Mm \in \mathbb{R}^{n \times m}$, $\|\Mm\|_F$ denotes the
Frobenius norm, $\|\Mm\|_*$ the nuclear norm, and
$\|\Mm\|_1$the entrywise $\ell_1$ norm.
We denote by $\operatorname{rank}(\Mm)$ the rank of $\Mm$ and by
$\|\Mm\|_0$ its number of nonzero entries. Given a set $S \subseteq [n]$, we write $\onev_S \in \{0,1\}^n$ for its
indicator vector and $|S|$ its cardinality. For an optimisation problem over a vector $\xv$,
we denote an optimal solution by $\xv^\star$. 
Throughout the paper, for a given integer-weighted directed graph, $n$ denotes the number of nodes,  $E$ denotes set of edges and $\Am \in \mathbb{Z}^{n \times n}$ denotes an
integer-weighted adjacency matrix. We write decompositions of the form
\begin{equation}
    \label{eq: main_decomposition}
    \Am = \Lm_0 + \Rm_0,
\end{equation}
where $\Lm_0$ is the structured component and $\Rm_0$ is the
residual. We use the notation $\Lm_0$ for the low-rank component in order to avoid confusion with the graph Laplacian.

\textbf{Sparse plus low-rank decompositions.}
Sparse plus low-rank decompositions model an observed matrix in the form \eqref{eq: main_decomposition}, where the two components satisfy $\operatorname{rank}(\Lm_0) \ll n$ and $\| \Rm_0\|_0 \ll n^2$.

The low-rank component is intended to capture global or latent structure,
whereas the sparse component captures localized deviations, corruptions, or
outliers. Classical convex relaxations replace the rank by the
nuclear norm and the number of nonzeros by the entrywise $\ell_1$ norm.

\textbf{Robust principal component analysis.}
RPCA is a canonical sparse plus low-rank model. Given an observed matrix $\Am$, RPCA seeks a decomposition (\ref{eq: main_decomposition})
by solving the convex optimisation problem
\begin{equation}
    \min_{\Lm_0, \Rm_0}
    \|\Lm_0\|_* + \lambda \|\Rm_0\|_1
    \quad
    \text{subject to}
    \quad
    \Am = \Lm_0 + \Rm_0,
    \label{eq:rpca}
\end{equation}
where $\lambda > 0$ controls the low-rank--sparsity tradeoff \citep{RPCA}. Under suitable incoherence and sparsity assumptions, RPCA can exactly recover low-rank matrices with sparse corruptions. However, these assumptions are poorly matched to sparse graph adjacency matrices (see Appendix \ref{app:slr}).
This motivates decompositions in which the structured component is not only
low-rank but also sparse. 

\textbf{Sparse principal component analysis.}
SPCA modifies PCA by imposing sparsity on the principal directions. 
A typical rank-$K$ formulation estimates sparse vectors $\Vm_1,\ldots,\Vm_K$ through
\begin{equation}
    \min_{\Um,\Vm}
    \|\Am - \Um\Vm^\top\|_F^2
    + \lambda \sum_{k=1}^K \|\Vm_k\|_1,
    \qquad
    \text{subject to suitable normalization constraints.}
    \label{eq:spca}
\end{equation}
Different variants use different normalizations, deflation schemes, and sparsity penalties \citep{SCoTLASS2003, SPCA2006}.
Although SPCA promotes sparse low-rank factors, it is not directly adapted to adjacency matrices since the factors are real-valued rather than discrete vertex indicators, and factor sparsity does not ensure that the residual $\Rm_0$ remains sparse. 
Thus, SPCA does not, by itself, provide the sparse, combinatorial, and storage-efficient graph decomposition required here.

\section{Signed Biclique Decomposition}
\label{sec:sbd}

We introduce SBD, a discrete low-rank model that represents an integer-weighted directed adjacency matrix as a sum of signed biclique blocks plus a sparse residual.

\subsection{Model}
\label{sec:sbd_model}

Let $\Am \in \mathbb Z^{n \times n}$ be a directed integer-weighted adjacency matrix. For each component $t$, let
$S_t,T_t \subseteq [n]$ denote source and target node sets, and let
$\alpha_t \in \{-1,+1\}$ denote its sign. We write $\wv_t = \mathbf 1_{S_t}, \: 
    \qv_t = \mathbf 1_{T_t},$
where $\wv_t,\qv_t \in \{0,1\}^n$ are binary indicator vectors.
A signed biclique is the rank-one integer matrix
$\alpha_t \wv_t \qv_t^\top$. Given $K$ components, SBD writes
\begin{equation}
    \label{eq:sbd_decomposition}
    \Am
    =
    \mathbf L_0^{(K)} + \Rm_0^{(K)}
    =
    \sum_{t=0}^{K-1}
    \alpha_t \wv_t \qv_t^\top
    +
    \Rm_0^{(K)} .
\end{equation}
Since each biclique is rank one, $\operatorname{rank}(\mathbf L_0^{(K)}) \le K$.
The decomposition has two goals. First, the biclique component $\Lm_0^{(K)}$ should capture the dominant 
structure of $\Am$, leaving a small residual $\Rm_0^{(K)}$, \textit{i.e.}, $\|\Rm_0^{(K)}\|_F^2 \ll \|\Am\|_F^2$. Second, the representation should be compact.
We encode each component by its supports and its sign, giving a storage cost of $|S_t|+|T_t|+1$. The residual is stored sparsely. This motivates the cost $
    \sum_{t=0}^{K-1} (|S_t|+|T_t|+1)
    +
    \|\Rm_0^{(K)}\|_0$, and the corresponding compression factor
\begin{equation}
    \label{eq:sbd_cf}
    \operatorname{CF}
    =
    \frac{\|\Am\|_0}
    {
    \sum_{t=0}^{K-1} (|S_t|+|T_t|+1)
    +
    \|\Rm_0^{(K)}\|_0
    } .
\end{equation}
Thus, larger values of $\operatorname{CF}$ correspond to more compact 
representations of the input graph and directly reduce the number of 
arithmetic operations required for sparse matrix computations.

\subsection{Fixed-rank SBD objective}
\label{sec:sbd_fixed_rank}

For a fixed number of components $K$, we search for an SBD as the following discrete regularized optimisation problem.

\textbf{Fixed-rank signed biclique approximation.}
Given $\lambda \ge 0$ and $K \in \mathbb N$, solve
\begin{equation}
    \label{eq:sbd_fixed_rank_objective}
    \min_{\substack{
        \wv_t,\qv_t \in \{0,1\}^n \\
        \alpha_t \in \{-1,+1\}
    }}
    \left\|
        \Am
        -
        \sum_{t=0}^{K-1}
        \alpha_t \wv_t \qv_t^\top
    \right\|_F^2
    +
    \lambda
    \sum_{t=0}^{K-1}
    \left(
        \|\wv_t\|_2^2
        +
        \|\qv_t\|_2^2
    \right).
\end{equation}

The first term controls reconstruction error:
\[
    \left\|
        \Am - \mathbf L_0^{(K)}
    \right\|_F^2
    =
    \|\Rm_0^{(K)}\|_F^2 .
\]
The second term controls the length of the biclique component.
Indeed, since $\wv_t$ and $\qv_t$ are binary, $ \|\wv_t\|_2^2 = |S_t|$ and$\: \|\qv_t\|_2^2 = |T_t|.$ Thus, $\lambda$ trades off reconstruction accuracy against biclique size.
Small values of $\lambda$ favor aggressive reconstruction, while larger values favor smaller and denser components (see Appendix \ref{app:lambda}).
The Frobenius loss in \eqref{eq:sbd_fixed_rank_objective} also promotes sparse residuals. Since
$\Am$ and all signed bicliques are integer-valued, the residual
$\Rm_0^{(K)}$ is integer-valued. Therefore, $\|\Rm_0^{(K)}\|_0\le \|\Rm_0^{(K)}\|_F^2$ and minimizing the squared residual hence upper-bounds the number of residual entries that must be stored explicitly.

\subsection{Greedy SBD objective}
\label{sec:sbd_greedy}

The problem in~\eqref{eq:sbd_fixed_rank_objective} is a combinatorial optimisation
problem over $K$ coupled bicliques. We therefore use a greedy approximation
that iteratively extracts one signed biclique at a time, in the spirit of
matching pursuit~\citep{Mallat1993}, but with data-dependent combinatorial
atoms.
Let $\Rm_0^{(0)}=\Am$. At iteration $t$, the next component is
chosen by solving the rank-one problem
\begin{equation}
    \label{eq:sbd_greedy_objective}
    (\alpha_t^\star,\wv_t^\star,\qv_t^\star)
    =
    \argmin_{\substack{
        \wv,\qv \in \{0,1\}^n \\
        \alpha \in \{-1,+1\}
    }}
    \left\|
        \Rm_0^{(t)}
        -
        \alpha \wv \qv^\top
    \right\|_F^2
    +
    \lambda
    \left(
        \|\wv\|_2^2
        +
        \|\qv\|_2^2
    \right).
\end{equation}
The residual is then updated as $\Rm_0^{(t+1)}
    =
    \Rm_0^{(t)}
    -
    \alpha_t^\star
    \wv_t^\star
    (\qv_t^\star)^\top.$ An exact solver stops when no nonempty biclique improves the objective. Our finite seed search (Section~\ref{sec:sbd_algorithms}) stops when its search budget is exhausted or no returned candidate has positive gain. This provides guarantees on the properties of the remainder obtained using either the exact solver or our heuristic methodology (Theorem~\ref{thm:sbd_main}).

\subsection{Guarantees}
\label{sec:sbd_guarantees}

The following result establishes the theoretical guarantees of greedy minimizing in~\eqref{eq:sbd_greedy_objective}. Full proof is deferred to Appendix~\ref{app: proof_main}.

\begin{definition}
    We call \textit{gain} the following function:
    \begin{equation}
    \label{eq:gain}
    G_\lambda(\alpha,S,T,\Rm_0)
  = 2\alpha\sum_{(i, j) \in S \times T}\Rm_0(i, j) - |S| |T| - \lambda (|S| + |T|).
    \end{equation}
\end{definition}

Note that minimizing \eqref{eq:sbd_greedy_objective} is equivalent to maximizing $G_\lambda$.

\begin{theorem}
\label{thm:sbd_main}
Let $\lambda\ge0$. Start with $\Rm_0^{(0)}=\Am$. At each accepted step update
$\Rm_0^{(t+1)}=\Rm_0^{(t)}-\alpha_t\onev_{S_t}\onev_{T_t}^{\top}$,
where $G_t=G_\lambda(\alpha_t,S_t,T_t,\Rm_0^{(t)})>0$. Then:
\begin{enumerate}[leftmargin=0.5cm]
\item \textbf{Residual decrease.} Each step satisfies
\begin{equation}\label{eq:step_decrease}
\|\Rm_0^{(t+1)}\|_F^2
=\|\Rm_0^{(t)}\|_F^2-G_t-\lambda(|S_t|+|T_t|).
\end{equation}
There are at most $K\le\|\Am\|_F^2$ steps. If $\lambda>0$, also
$K\le\lfloor\|\Am\|_F^2/(2\lambda)\rfloor$.
\item \textbf{Constraints at stopping.} Suppose that stopping certifies $G_\lambda\le0$ for every candidate in a set $\mathcal C_K$, evaluated on the final residual. For each $(\alpha,S,T)\in\mathcal C_K$,
\begin{equation}\label{eq:terminal_density}
\alpha d_{\Rm_0^{(K)}}(S,T)
\le\frac12+\frac{\lambda}{2}\left(\frac1{|S|}+\frac1{|T|}\right),
\end{equation}
where $d_R(S,T)=\frac{\sum_{i\in S,j\in T}R(i,j)}{|S||T|}$.
If both signs were checked for the same supports, this also bounds $|d_{\Rm_0^{(K)}}(S,T)|$.
\item \textbf{Compression bound.} Define
$J_t=\|\Rm_0^{(t)}\|_F^2+\lambda\sum_{\ell<t}(|S_\ell|+|T_\ell|)$.
 $J_{t+1} =J_t -G_t$ and $J_t$ is non increasing. For $\lambda\ge1$,
\begin{equation}\label{eq:cf_upper_bound}
\mathrm{CF}\ge\frac{\|\Am\|_0}{J_K + K}.
\end{equation}
Also, $J_t + t$ is non increasing for $\lambda \in \mathbb N^*$. It shows that our update increases a surrogate lower bound on CF at each step.
\end{enumerate}
\end{theorem}

\begin{corollary}\label{cor:sbd_main}
If the stopping certificate includes every singleton support with both signs, then
$|\Rm_0^{(K)}(i,j)|\le\lfloor(1+2\lambda)/2\rfloor$ for all $i,j$.
Thus $\Rm_0^{(K)}=0$ for $\lambda<1/2$, and its entries have magnitude at most one for $\lambda\in[1/2,3/2)$.
\end{corollary}

A sequential search that checks positive gain on the exact current residual before acceptance satisfies the descent assumptions of Theorem~\ref{thm:sbd_main}. This includes both SBDGreedy and SBDFast. Neither method needs an exact
rank-one solver for this descent guarantee.

The stopping constraints apply only to candidates checked on the final residual. A global rank-one oracle checks all candidates and implies Corollary~\ref{cor:sbd_main}. A finite seed search does not provide that global certificate. Appendix~\ref{app:local_guarantees} gives additional theoretical guarantees. Appendix~\ref{app:ilp_comparison} compares our search with a rank-one solver on small graphs.

\subsection{Algorithms}
\label{sec:sbd_algorithms}

SBDGreedy uses alternating optimisation (AO) to maximise~\eqref{eq:gain}. For a fixed sign and row support, columns contribute independently. With a density floor $d_{\min}$, the update is
\[
 T\leftarrow\{j\in V:2\alpha\sum_{i\in S}\Rm_0(i,j)
 >\max(|S|+\lambda,2d_{\min}|S|)\}.
\]
The row update is symmetric. The density floor limits holes in the selected block. Appendix~\ref{app:impl_details} derives the updates and local candidate sets.
SBDGreedy checks the full gain before each sequential extraction, so Theorem~\ref{thm:sbd_main} applies. Its stopping certificate covers only the candidates checked.

\textbf{SBDFast} searches seed batches in parallel and applies proposals sequentially. Before each acceptance, it recomputes the gain on the current exact residual, including updates from earlier proposals in the batch. The search first proposes positive blocks, then searches both signs. See Appendix~\ref{app:fast_tricks} for more details.

\begin{wrapfigure}{r}{0.55\textwidth}
\centering
\vspace{-10pt}

\small
\hrule height 0.8pt
\captionsetup{skip=3pt}
\captionof{algorithm}{SBDGreedy: signed biclique extraction}
\label{alg:sbd}
\hrule height 0.4pt
\vspace{3pt}

\begin{algorithmic}[1]
\REQUIRE $\Am$, $\lambda$, $d_{\min}$,
         seed budget $M$ per sign,
         AO cap $J$, extraction cap $K_{\max}$

\STATE $\Rm_0 \leftarrow \Am$

\REPEAT
    \STATE For each sign $\alpha$, take up to $M$ rows
    ranked by $\sum_j \max(\alpha \Rm_0(i,j),0)$

    \FORALL{signed seeds $(u,\alpha)$ \textbf{in parallel}}
        \STATE Form local candidate sets $(U,V)$ around $u$;
        set $S \leftarrow \{u\}$, $T \leftarrow V$
        \STATE Alternate for at most $J$ iterations
        \STATE Stop AO if both supports are unchanged
        or either is empty
    \ENDFOR

    \STATE Select the nonempty candidate
    $(\alpha^\star,S^\star,T^\star)$
    with largest full gain

    \IF{no candidate has positive gain}
        \STATE \textbf{stop}
    \ENDIF

    \STATE $\Rm_0 \leftarrow
    \Rm_0-\alpha^\star
    \mathbf{1}_{S^\star}
    \mathbf{1}_{T^\star}^{\top}$

\UNTIL{at most $K_{\max}$ blocks have been extracted}

\end{algorithmic}
\vspace{3pt}
\hrule height 0.8pt

\vspace{-10pt}
\end{wrapfigure}
\subsection{Accelerating SpMM with SBD}

Standard truncated SVD (\(k\)-SVD) accelerates dense matrix multiplication by replacing a full matrix with dense low-rank factors, but this is poorly suited to graph SpMM: the dense singular vectors destroy the sparsity structure that makes SpMM efficient. In contrast, SBD accelerates SpMM while preserving sparsity and remaining lossless. After \(k\) decomposition steps, defining 
$\wv = (\wv_0,\ldots,\wv_{k-1}),
\mathbf{\Lambda} = \mathrm{diag}(\alpha_0,\ldots,\alpha_{k-1}), 
\qv = (\qv_0,\ldots,\qv_{k-1})$, we obtain
\[
\begin{aligned}
\Am &= \sum_{t=0}^{k-1}\alpha_t\wv_t\qv_t^\top
       + \Rm_0^{(k)} \\
    &= \wv\mathbf{\Lambda}\qv^\top + \Rm_0^{(k)}.
\end{aligned}
\]
This factorization yields a sparser representation of the propagation operator, thereby reducing the number of multiplications in SpMM. Computationally, it replaces a single SpMM by two sequential sparse multiplications involving thinner factors. As a result, the method reduces arithmetic and communication costs at the expense of decreased parallelism.
In practice, the decomposition can be implemented efficiently in CUDA, where the reduced message volume helps amortize kernel-launch overhead. 

\section{Experiments}
\label{sec:experiments}

We evaluate SBD along three axes. First, we test whether the signed biclique model can recover sparse low-rank graph structure in a controlled denoising setting. Second, we compare the scalability of SBD against existing sparse low-rank matrix decomposition methods on real graphs (see Appendix~\ref{app:dataset}). Third, we evaluate whether the learned representation yields practical speedups for SpMM, the core linear-algebraic primitive in many graph representation learning methods like graph neural networks \cite{kipf2017semi}.

\subsection{Experimental setup}
\label{sec:exp_setup}

\textbf{Graph denoising.} We follow the benchmark of \citet{Kanada2018}. The clean graph has ten cliques of 20 nodes ($p_{\mathrm{in}}=1$). We add inter-community edges with probability $q\in\{0.05,0.15,0.25,0.35\}$. We threshold $\Lm_0$ at 0.5 and measure F$_1$ against the clean graph, excluding the diagonal. We compare SBDGreedy with RPCA~\citep{RPCA}, sparse SVD, Kanada et al.~\citep{Kanada2018}, and Savalle et al.~\citep{Savalle2012EstimationOS}, solved by ADMM. Each method uses 30 Optuna trials on five tuning graphs per noise level. Evaluation uses ten separate graphs. Appendix~\ref{app:hyperparameters} gives all ranges and fixed settings.

\textbf{Decomposition scalability.} We measure decomposition time on real graphs with fixed settings. SBDGreedy and SBDFast run on CPU using eight threads, while the continuous baselines run on an RTX~3090 GPU. These settings do not match reconstruction quality. For compression, we select the SBDFast decomposition with the highest CF from 30 Optuna trials per graph.

\textbf{SpMM acceleration.} We reuse the selected decomposition across feature widths and GPUs. We compare with cuSPARSE, Sputnik~\citep{sgk_sc2020}, dgSPARSE, GE-SpMM~\citep{9355302}, and CBM~\citep{Joao2025}. SBD\_base uses our kernel without bicliques ($\Lm_0=0$ and $\Rm_0=\Am$). All kernels are measured on RTX~3090, A40, and L40S, with ten warm-up and 50 timed launches; we report median latency. We use float32 features with $F\in\{32,128,256\}$ columns. Multiplication uses $\Lm_0+\Rm_0=\Am$, retaining the exact remainder. Appendix~\ref{app:kernel} describes the kernel and Appendix~\ref{app:hyperparameters} gives the timing protocol.

\subsection{Denoising sparse low-rank graph structure}
\label{sec:denoising}

\begin{wraptable}{r}{0.55\linewidth}

\centering\small
\caption{SBM denoising F$_1$ ($\uparrow$), $K=10$, $N=200$, $p_{\mathrm{in}}=1$. Means over ten evaluation graphs.}
\label{tab:denoising}
\begin{tabular*}{\linewidth}{@{\extracolsep{\fill}}lcccc@{}}
\toprule
{Method} & {0.05} & {0.15} & {0.25} & {0.35} \\
\midrule
{Noisy input} & {0.808} & {0.585} & {0.457} & {0.375} \\
{Savalle} & {1.000} & {0.989} & {0.869} & {0.744} \\
{RPCA} & {1.000} & {1.000} & {0.999} & {0.707} \\
{Sparse SVD} & {1.000} & {0.999} & {0.970} & {0.749} \\
{Kanada} & {1.000} & {1.000} & {1.000} & {0.992} \\
{SBDGreedy (ours)} & {1.000} & {1.000} & {0.999} & {0.992} \\
\bottomrule
\end{tabular*}

\end{wraptable}

Table~\ref{tab:denoising} reports denoising performance across corruption levels. SBD exactly recovers the clean graph on the evaluation seeds in the low-noise regimes $q\leq0.15$ and remains competitive as the inter-community noise increases. At $q=0.25$, SBD substantially improves over the noisy baseline and Savalle et al., and remains close to the best-performing methods. At the highest noise level, SBD and Kanada et al. both score 0.992, while SBD outperforms RPCA and sparse SVD. These results indicate that SBD captures the intended sparse block structure while using discrete biclique factors rather than dense or continuous low-rank components.

\subsection{Scalability of the decomposition}
\label{sec:decomp_scalability}

\begin{table}[b]
\centering\small
\caption{Runtime in seconds under a 60-second compute limit. OOT: time limit; OOM: allocation failure or dense input exceeding the 24 GiB budget. SBD uses eight CPU threads; continuous baselines use RTX~3090.}
\label{tab:real_graph_runtime}

\begin{tabular*}{\linewidth}{@{\extracolsep{\fill}}lrrrrr@{}}
\toprule
Method & Cora & CiteSeer & ca-HepPh & ca-AstroPh & COLLAB \\
\midrule
SBDGreedy & 5.99 & 5.59 & OOT & OOT & OOT \\
SBDFast & 0.022 & 0.026 & 0.35 & 0.60 & 12.3 \\
Kanada & OOT & OOT & OOT & OOM & OOM \\
RPCA & 4.36 & 7.10 & OOT & OOT & OOM \\
Sparse SVD & 6.30 & 2.08 & 55.0 & OOT & OOM \\
Savalle & OOT & OOT & OOT & OOM & OOM \\
\bottomrule
\end{tabular*}
\end{table}

Table~\ref{tab:real_graph_runtime} compares decomposition runtimes under a $60$ second budget. Kanada et al. and Savalle reach the time limit on Cora, CiteSeer and ca-HepPh, and run out of memory on ca-AstroPh. Savalle uses the same symmetry-based ADMM solver as in denoising. RPCA succeeds on Cora and CiteSeer, while its dense intermediate matrices scale as $\mathcal O(n^2)$ in memory. Sparse SVD is computationally efficient, but its continuous factors are not binary indicators and therefore do not provide the signed biclique representation needed for our SpMM kernel; see Appendix~\ref{app:slr}. In contrast, SBDFast completes all five graphs on CPU, from 0.022 seconds on Cora to 12.3 seconds on COLLAB, and SBDGreedy remains practical on the smaller instances. Continuous baselines are not evaluated on COLLAB because its dense input exceeds the memory limit.

\subsection{Representation quality}
\label{sec:representation_quality}

\begin{table}[h]
\caption{SBDFast decomposition quality using the highest-CF trial per graph. CF counts stored entries; CR counts bytes relative to CSR. SBD uses its kernel arrays; CBM uses the storage model defined by its authors.}
\label{tab:decomp_quality}

\centering\small\setlength{\tabcolsep}{4pt}
\begin{tabular*}{\linewidth}{@{\extracolsep{\fill}}lrrrrrr@{}}
\toprule
Dataset & $N$ & $\|\Am\|_0$ & $K$ & CF & CR$_{\mathrm{SBD}}$ & CR$_{\mathrm{CBM}}$ \\
\midrule
Cora & 2,708 & 10,556 & 181 & 1.05 & 0.99 & 1.06 \\
PubMed & 19,717 & 88,648 & 1,651 & 1.08 & 1.04 & 1.05 \\
ca-HepPh & 12,008 & 237,010 & 2,437 & 3.23 & 3.43 & 2.75 \\
ca-AstroPh & 18,772 & 396,160 & 10,519 & 2.15 & 2.70 & 1.73 \\
coPapersDBLP & 540,486 & 30,491,458 & 86,855 & 8.57 & 10.3 & 5.98 \\
coPapersCiteseer & 434,102 & 32,073,440 & 84,862 & 11.3 & 14.4 & 9.91 \\
COLLAB & 372,474 & 24,572,158 & 53,646 & 10.9 & 12.2 & 11.0 \\
\bottomrule
\end{tabular*}

\end{table}

Table~\ref{tab:decomp_quality} reports the size of the SBDFast representation on larger real graphs. We distinguish between the compression factor CF, which counts support memberships, signs and residual entries, and the memory compression ratio CR, which compares representation bytes with CSR. CBM uses a graph compression format based on ordered neighborhood differences. SBD optimizes a different objective: it factorizes shared neighborhoods into signed biclique components while preserving compatibility with ordinary linear algebra. Thus, the relevant question is not only whether SBD compresses the graph, but whether the resulting operator representation accelerates SpMM. Appendix~\ref{app:storage_bytes} gives both storage models.

The results show two regimes. On graphs with limited shared-neighborhood structure, such as Cora and PubMed, SBD is close to break-even in CF and does not provide substantial memory compression. Their byte ratios are 0.99 and 1.04. On graphs with stronger shared-neighborhood structure, such as coPapersDBLP, coPapersCiteseer and COLLAB, SBD obtains compression factors of 8.57, 11.3 and 10.9. Under the stated byte models, CBM compresses slightly better on Cora and PubMed, while SBD compresses better on the other five graphs.

\subsection{SpMM acceleration}
\label{sec:spmm_results}

\begin{figure}[t]
\centering
\includegraphics[width=\linewidth]{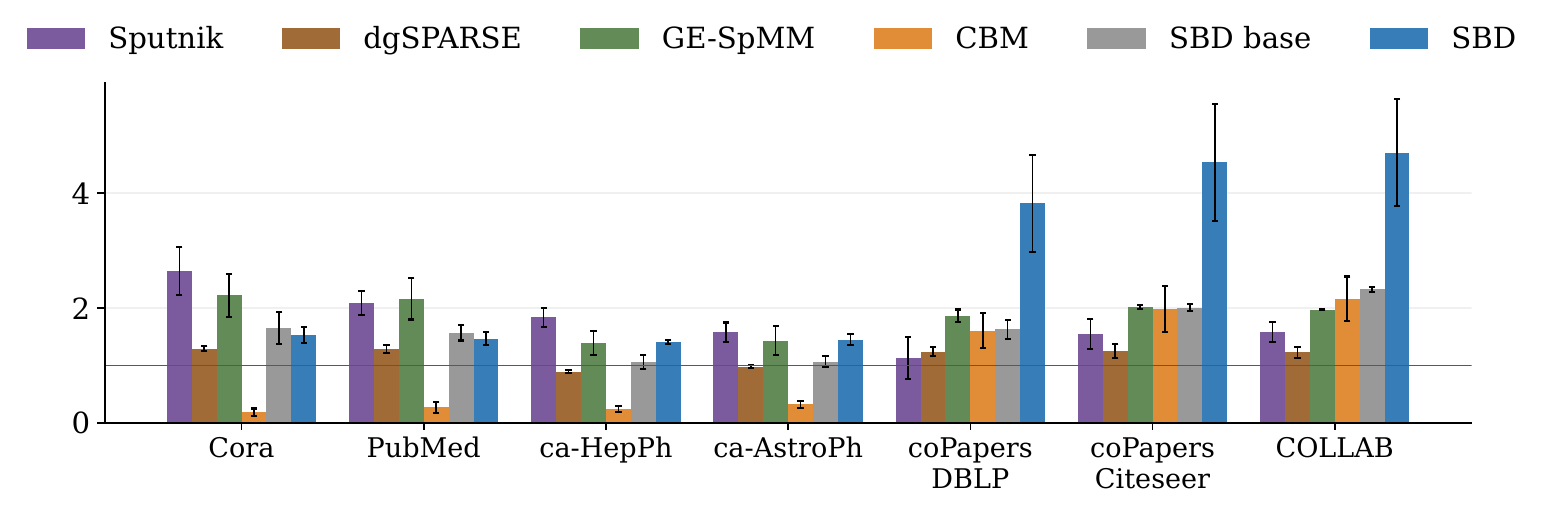}
\caption{SpMM speedup over cuSPARSE, averaged over $F\in\{32,128,256\}$ within each GPU and then over RTX~3090, A40 and L40S. Error bars show variation across GPU averages. SBD\_base uses no bicliques. Per-width results are in Appendix~\ref{app:kernel_comparison_values}.}
\label{fig:kernel_comparison}
\end{figure}

Figure~\ref{fig:kernel_comparison} reports SpMM speedups relative to cuSPARSE. Averaged across the three GPUs, SBD is faster than cuSPARSE on every tested graph at each feature width, with the largest gains on high-CF datasets. On COLLAB, SBD reaches $6.37\times$ speedup at $F=32$ and remains above $3.8\times$ at $F=128$ and $F=256$. On coPapersCiteseer, SBD obtains $5.80\times$ speedup at $F=32$ and above $3.8\times$ at larger feature dimensions. These results show that the signed biclique representation can reduce repeated computation while preserving the exact SpMM operator.

Figure~\ref{fig:kernel_comparison} includes CBM alongside the GPU SpMM kernels. Compared with CBM, SBD is faster on every dataset and feature dimension in the GPU averages of Table~\ref{tab:speedup_sbd}. CBM benefits from compression on the largest community graphs, but is slower than cuSPARSE on the smaller graphs. SBD is also faster than Sputnik, dgSPARSE and GE-SpMM on the three largest graphs. However, Sputnik is faster on Cora, PubMed and ca-HepPh at every tested width. 

SBD\_base uses the same kernel without bicliques. Comparing it with SBD isolates the effect of the decomposition. On COLLAB at $F=32$, it gives $1.75\times$ speedup compared with SBD's $6.37\times$, whereas it is faster than SBD on PubMed. The two-pass biclique computation amortizes column aggregation across rows sharing the same target set, but its overhead outweighs these benefits on weakly compressible graphs. Additional kernel-level details are provided in Appendix~\ref{app:kernel}.

\begin{figure}[t]
\centering
\includegraphics[width=\linewidth]{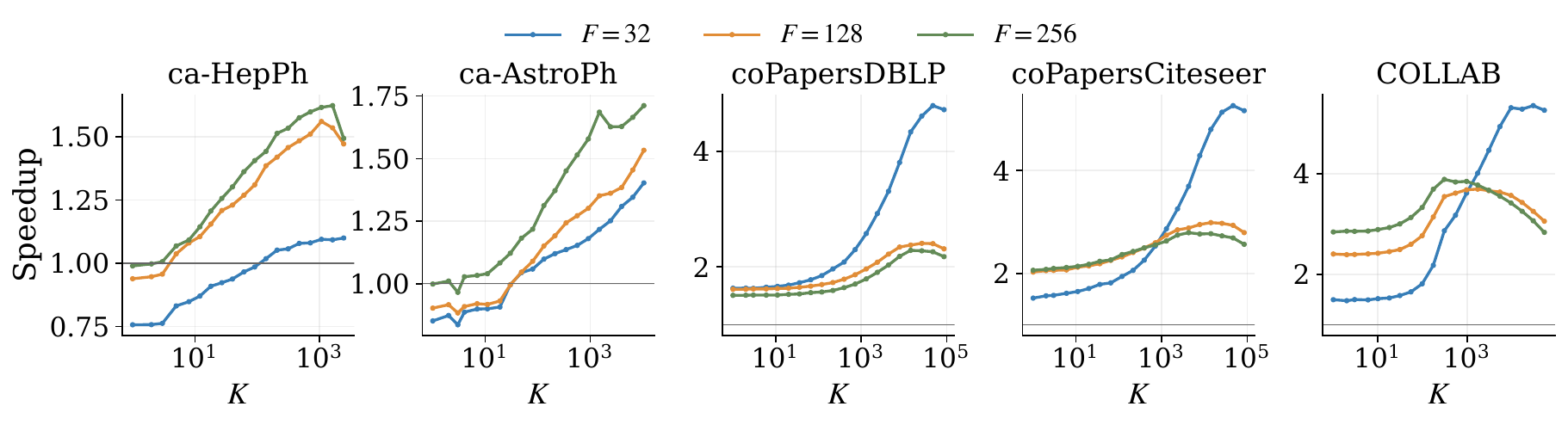}
\caption{SpMM speedup over PyTorch sparse multiplication vs.\ number of bicliques $K$ used on L40S, for $F\in\{32,128,256\}$. Bicliques are added in decreasing area order and the exact residual is rebuilt for each prefix.}
\label{fig:k_sweep_main}
\end{figure}

\subsection{When does SBD help?}
\label{sec:spmm_analysis}

The largest speedups in Figure~\ref{fig:kernel_comparison} occur on the graphs with the highest compression factors in Table~\ref{tab:decomp_quality}. The observed speedup is smaller than CF on these graphs because SBD replaces a single CSR SpMM by a two-pass computation involving the biclique factors and the residual. CF counts entries rather than GPU costs, so it does not predict latency directly. Graphs such as Cora and PubMed provide little compression, and the additional two-pass overhead can dominate the benefit of the blocks.

Figure~\ref{fig:k_sweep_main} studies the effect of using only the top $K$ bicliques, ordered by decreasing area. Increasing $K$ moves more of the graph from the sparse residual into the structured two-pass path, but the benefit depends on the feature dimension. At $F=32$, speedup continues to improve over a large range of $K$: COLLAB reaches its best sampled speed using about $56\%$ of the available bicliques, and coPapersCiteseer behaves similarly. At $F=128$ and $F=256$, these graphs reach their best sampled speed earlier. On COLLAB, the corresponding fractions are about $3.2\%$ and $0.6\%$. 

This is not a universal cutoff: ca-AstroPh reaches its best sampled speed at the full decomposition for all three widths. Thus, the decomposition is most beneficial when the graph contains many reusable shared-neighborhood blocks, and using all blocks need not give the fastest multiplication. Appendix~\ref{app:k_sweep_analysis} discusses the feature-width dependence. Both experiments use PyTorch sparse CSR multiplication with CUDA cuSPARSE backend as the reference.

\subsection{Limitations}
\label{sec:limitations}

Our approach relies on structural redundancy in the graph for effective representation. Graphs with few shared-neighborhood structures require more biclique components and reduce compression efficiency.  Signed binary factors may be less effective for weighted graphs (varying edge weights can require more components). Finally, our decomposition does not always lead to faster SpMM, as the overhead may outweigh its benefits on graphs with limited redundancy.

\section{Conclusion}

We introduced signed biclique decomposition (SBD), a sparse plus 
low-rank-and-sparse decomposition for integer-weighted graphs. Unlike 
classical low-rank methods such as RPCA, SBD represents the structured 
component as a sum of sparse signed binary rank-one factors, preserving 
the discrete, combinatorial, and computational properties of the graph. 
We formulated fixed-rank and greedy SBD objectives with theoretical guarantees. Empirically, SBD recovers latent block structure in noisy graphs competitively with RPCA-based methods, while remaining scalable to large graphs. On graphs with strong shared-neighborhood structure, 
SBD achieves compression factors above $10$ and SpMM speedups up to $6\times$ over cuSPARSE, demonstrating that sparse combinatorial 
decompositions can serve as a practical and interpretable alternative 
to dense low-rank representations for graph data.

\section*{Acknowledgment}

The authors declare no competing interests. The authors acknowledge the French National Research Agency (ANR) for the JCJC project DeSNAP ANR-24-CE23-1895-01. This work is also
supported by Hi! PARIS,  ANR/France 2030 program
ANR-23-IACL-0005, ANR-23-CMAS-0033, and ANR-23-RHUS-0002 (Innov4-ePiK). 
\bibliographystyle{plainnat}
\bibliography{references}

\newpage
\appendix

\doparttoc 
\faketableofcontents 

\addcontentsline{toc}{section}{Appendix} 
\part{{\Large Appendix}} 
\parttoc 
\newpage

\section*{Table of Abbreviations}
\addcontentsline{toc}{section}{Table of Abbreviations}
\label{sec:abbreviations}

To assist the reader, Table~\ref{tab:abbreviations} provides a summary of the main abbreviations and acronyms used throughout this paper.

\begin{table}[H]
\centering
\caption{{Summary of abbreviations and acronyms used in this paper.}}
\small

\begin{tabular}{ll}
\toprule
{\textbf{Abbreviation}} & {\textbf{Description}} \\
\midrule
{SLR} & {Sparse plus Low-Rank (matrix decomposition model)} \\
{SLRS} & {Sparse plus Low-Rank-and-Sparse (our graph-adapted model)} \\
{RPCA} & {Robust Principal Component Analysis} \\
{SPCA} & {Sparse Principal Component Analysis} \\
{SBD} & {Signed Biclique Decomposition (proposed framework)} \\
{SpMM} & {Sparse Matrix-Matrix Multiplication ($\Am \times \Xm$)} \\
{CF} & {Compression Factor ($\|\Am\|_0 / \mathrm{repr\_cost}$)} \\
{CR} & {Memory Compression Ratio ($\mathrm{MB}(\mathrm{CSR}) / \mathrm{MB}(\mathrm{format})$)} \\
{AO} & {Alternating Optimization} \\
{ILP} & {Integer Linear Programming (exact combinatorial solver)} \\
{CSR} & {Compressed Sparse Row format} \\
{SBM} & {Stochastic Block Model} \\
\bottomrule
\end{tabular}

\label{tab:abbreviations}
\end{table}

\clearpage
\section{Proofs}
\label{app:proofs}

\label{sec:appendix_proofs_algos}

\subsection{{Proof of Theorem \ref{thm:sbd_main}}}
\label{app: proof_main}

\begin{proof}[Proof of Theorem~\ref{thm:sbd_main}]
{\textbf{(1) Residual decrease and finite termination.}
Write $R=\Rm_0^{(t)}$} and {$B=\alpha_t\onev_{S_t}\onev_{T_t}^{\top}$. Expanding the squared norm gives
\[
\|R-B\|_F^2=\|R\|_F^2-2\alpha_t\sum_{i\in S_t,j\in T_t}R(i,j)+|S_t||T_t|.
\]
Substituting the gain gives~\eqref{eq:step_decrease}. Each accepted step strictly decreases the squared norm. This norm is a nonnegative integer, so there are at most $\|\Am\|_F^2$ steps.
For} $\lambda>0$, {nonempty supports give $|S_t|+|T_t|\ge2$. Summing the decreases over $K>0$ accepted steps yields}
\begin{equation}\label{eq:residual_decrease}
0\le\|\Rm_0^{(K)}\|_F^2<\|\Am\|_F^2-2\lambda K.
\end{equation}
{Hence $K\le\lfloor\|\Am\|_F^2/(2\lambda)\rfloor$. The bound also holds for $K=0$.

\textbf{(2) Constraints at stopping.}
Let $R=\Rm_0^{(K)}$. A nonempty candidate $(\alpha,S,T)\in\mathcal C_K$ satisfies
\[
2\alpha\sum_{i\in S,j\in T}R(i,j)-|S||T|-\lambda(|S|+|T|)\le0.
\]
Dividing by $2|S||T|$ gives~\eqref{eq:terminal_density}. Checking both signs on the same supports bounds $|d_R(S,T)|$. The bound covers only candidates checked on this final residual.

\textbf{(3) Compression factor.}
Every nonzero integer has squared magnitude at least one. Thus}
\begin{equation}\label{eq:nonzero_bound}
\|\Rm_0^{(K)}\|_0\le\|\Rm_0^{(K)}\|_F^2.
\end{equation}
{For $\lambda\ge1$}, {the representation cost satisfies}
\begin{align*}
C_K&=\|\Rm_0^{(K)}\|_0+\sum_{t<K}(|S_t|+|T_t|)+K\\
&\le\|\Rm_0^{(K)}\|_F^2+\lambda\sum_{t<K}(|S_t|+|T_t|)+K=J_K+K.
\end{align*}
{Since $\Am\ne0$}, {dividing $\|\Am\|_0$ by these costs gives~\eqref{eq:cf_upper_bound}.
Equation~\eqref{eq:step_decrease} also gives $J_{t+1}=J_t-G_t$. Hence
\[
(J_{t+1}+t+1)-(J_t+t)=1-G_t.
\]
For positive integer $\lambda$, an accepted gain is an integer with $G_t\ge1$. The right-hand side is nonpositive, so the lower bound on CF does not decrease.}
\end{proof}

\begin{proof}[Proof of Corollary~\ref{cor:sbd_main}]
{The certificate checks both signs on every singleton. Choose the sign of a nonzero $R(i,j)$, where $R=\Rm_0^{(K)}$. Its gain satisfies
\[
2|R(i,j)|-1-2\lambda\le0.
\]
Since the residual is integer-valued, $|R(i,j)|\le\lfloor(1+2\lambda)/2\rfloor$. Zero entries also satisfy this bound. It is zero} for {$\lambda<1/2$ and one} for {$\lambda\in[1/2,3/2)$}.
\end{proof}

\clearpage
\section{{Theoretical analysis}}
\label{app:local_guarantees}

\subsection{{Additional Analysis on SBD Objective}}
\label{sec:problem2_details}
\paragraph{Expansion of the greedy SBD objective.}
Expanding the Frobenius norm in \eqref{eq:sbd_greedy_objective} and using 
the fact that $\|\wv_t \qv_t^\top\|_F^2 = \|\wv_t\|_F^2\|\qv_t\|_F^2 = 
|S_t||T_t|$ for binary vectors, the minimization reduces to the following 
(dropping the constant $\|\Rm_0^{(t)}\|_F^2$):

\begin{equation}
    \label{eq:problem2_expanded}
    \min_{\substack{S_t,\, T_t \\ \alpha_t \in \{-1,+1\}}}
    \; |S_t||T_t| - 2\alpha_t \sum_{(i,j) \in S_t \times T_t} \Rm_0^{(t)}(i,j)
    + \lambda\left(|S_t| + |T_t|\right).
\end{equation}

Recall that we define the block density of a matrix $\Mm$ on $S \times T$ as:

\begin{equation}
    \label{eq:block_density}
    d_{\Mm}(S, T) = \frac{1}{|S||T|}\sum_{(i,j) \in S \times T} \Mm(i,j).
\end{equation}

\paragraph{Substitution of the residual.}
Substituting $\Rm_0^{(t)} = \Am - \sum_{i=0}^{t-1}\alpha_i^*\wv^*_i(\qv^*_i)^\top$ 
into \eqref{eq:problem2_expanded} and using \eqref{eq:block_density} yields:

\begin{equation}
    \label{eq:problem2_t}
    \min_{\substack{S_t,\, T_t \\ \alpha_t \in \{-1,+1\}}}
    \; |S_t||T_t|\left(1 - 2\alpha_t\, d_{\Am}(S_t, T_t)\right)
    + 2\sum_{i=0}^{t-1}(\alpha_t\alpha_i^*)
    \left(\wv_t^\top \wv^*_i\right)\left(\qv_t^\top \qv^*_i\right)
    + \lambda\left(|S_t| + |T_t|\right).
\end{equation}

Since $\wv_t^\top \wv^*_i = |S_t \cap S_i^*|$ and 
$\qv_t^\top \qv^*_i = |T_t \cap T_i^*|$ for binary vectors, the cross term 
satisfies:

\begin{equation}
    \sum_{i=0}^{t-1}(\alpha_t\alpha_i^*)
    \left(\wv_t^\top \wv^*_i\right)\left(\qv_t^\top \qv^*_i\right)
    = |S_t||T_t|
    \sum_{i=0}^{t-1}(\alpha_t\alpha_i^*)
    \frac{|S_t \cap S_i^*|}{|S_t|}
    \frac{|T_t \cap T_i^*|}{|T_t|}.
\end{equation}

Factoring out $|S_t||T_t|$ from \eqref{eq:problem2_t}, the objective 
reduces to:

\begin{equation}
    \label{eq:problem2_factored}
    \min_{\substack{S_t,\, T_t \\ \alpha_t \in \{-1,+1\}}}
    \; |S_t||T_t| \cdot \mathcal{C}(S_t, T_t, \alpha_t),
\end{equation}

where:

\begin{equation}
    \label{eq:cost_per_area}
    \mathcal{C}(S_t, T_t, \alpha_t) 
    = 1 
    - 2\alpha_t\, d_{\Am}(S_t, T_t) 
    + 2\sum_{i=0}^{t-1}(\alpha_t\alpha_i^*)
    \frac{|S_t \cap S_i^*|}{|S_t|}
    \frac{|T_t \cap T_i^*|}{|T_t|}
    + \frac{\lambda}{|S_t|} + \frac{\lambda}{|T_t|}.
\end{equation}

\paragraph{Interpretation of $\mathcal{C}$.}
The three terms in \eqref{eq:cost_per_area} have distinct roles:

\begin{itemize}[leftmargin=0.5cm]
    \itemsep0em
    \item \textbf{Density reward} $-2\alpha_t\,d_{\Am}(S_t,T_t)$: 
    rewards bicliques that are aligned with the block structure of $\Am$. 
    {For nonnegative $\Am$}, {this term favours $\alpha_t=+1$ when the block density is positive. For signed input, it favours the sign of $d_{\Am}(S_t,T_t)$.}

    \item \textbf{Signed overlap penalty} 
    $2(\alpha_t\alpha_i^*)\frac{|S_t \cap S_i^*|}{|S_t|}
    \frac{|T_t \cap T_i^*|}{|T_t|}$: acts as an \emph{implicit memory 
    mechanism}. Same-sign overlap ($\alpha_t\alpha_i^* = +1$) raises 
    $\mathcal{C}$, discouraging the algorithm from covering the same 
    region twice with the same sign. Opposite-sign overlap 
    ($\alpha_t\alpha_i^* = -1$) lowers $\mathcal{C}$, encouraging 
    self-correction via an opposite-sign biclique on a previously 
    covered region.

    \item \textbf{Size penalty} $\lambda/|S_t| + \lambda/|T_t|$: 
     {tends to zero as both supports grow and equals $2\lambda$} for {a singleton, penalising} small {bicliques more strongly.} The parameter $\lambda$ directly 
    controls this bias.
\end{itemize}

We denote  \emph{overlap-corrected density} as $\tilde{d}_t(S_t, T_t) = d_{\Am}(S_t, T_t) - \sum_{i=0}^{t-1}\alpha_i^*\frac{|S_t \cap S_i^*|}{|S_t|} \frac{|T_t \cap T_i^*|}{|T_t|}$.

{$\mathcal{C}(S_t, T_t, \alpha_t) < 0$} implies:
\begin{equation}
  \alpha_t \tilde{d}_t(S_t, T_t) > \frac{1}{2}\left(1 + \frac{\lambda}{|S_t|} + \frac{\lambda}{|T_t|}\right).
\end{equation}
{Since the right-hand side is strictly positive, $\alpha_t^\star = \operatorname{sign}(\tilde{d}_t(S_t, T_t))$} and
\begin{equation}
  \label{eq:acceptance_condition}
  |\tilde{d}_t(S_t, T_t)| > \frac{1}{2}\left(1 + \frac{\lambda}{|S_t|} + \frac{\lambda}{|T_t|}\right).
\end{equation}

{Acceptance requires the absolute residual density to exceed $\frac{1}{2}(1 + \lambda/|S_t| + \lambda/|T_t|)$. Here $\tilde{d}_t(S_t, T_t) = d_{\mathbf{R}_0^{(t)}}(S_t, T_t)$} is the {current residual density. This has three consequences:}
(i) the threshold {is at least $\frac{1}{2}$; for $\lambda>0$, it} approaches {$\frac{1}{2}$ from above} for large blocks and {grows} for small {ones, so small blocks must be denser to be accepted;}
(ii) larger $\lambda$ {raises} the {threshold; in particular, when residual entries are bounded by $1$ in magnitude (e.g.\ binary $\mathbf{A}$ before any overlap), $|S_t| \times |T_t|$ block can be extracted, even when fully dense, only if $\lambda/|S_t| + \lambda/|T_t| < 1$};
(iii) same-sign past overlap reduces $\alpha_t \tilde{d}_t$, requiring higher 
raw density $d_{\Am}$ to justify revisiting an already-covered region.

\subsection{{Derivation of the AO closed-form update}}
\label{app:impl_details}

\paragraph{Separability of the gain at fixed $S$.}
Fix the sign $\alpha$ and the source set $S$.
Recall that the gain reads
\begin{equation}
  G_{\lambda}(\alpha, S,T, \Rm_0)
  \;=\;
  2\!\sum_{j\in T}\!\underbrace{\alpha\sum_{i\in S}\Rm_{0}(i,j)}_{\displaystyle c_j(S,\alpha)}
  \;-\; |S|\,|T| \;-\; \lambda\bigl(|S|+|T|\bigr),
\end{equation}
where {$c_j(S,\alpha)=\alpha\sum_{i\in S}\Rm_{0}(i,j)$}.
$G_{\lambda}$ decomposes as
\begin{equation}
  \label{eq:gain_decomp}
  G_{\lambda}(\alpha, S,T,\Rm_0)
  =
  \sum_{j\in T}\!\bigl(2\, c_j(S) - |S| - \lambda\bigr)
  \;-\;\lambda|S|.
\end{equation}
Each $j$ contributes independently: including $j$ in $T$ improves the gain
if and only if
\begin{equation}
  \label{eq:halfstep_deriv}
  \frac{ c_j(S,\alpha)}{|S|}
  =
  \frac{1}{|S|}\sum_{i\in S}\alpha\,\Rm_{0}(i,j)
  \;>\;
  \frac{1}{2} + \frac{\lambda}{2|S|}.
\end{equation}

The optimal $T$ at fixed $S$ {therefore} consists exactly of all columns whose
average connection to $S$ exceeds
$\tfrac{1}{2}+\tfrac{\lambda}{2|S|}${.
Denoting this set by}
\begin{equation}
  T^\star
  =
  \left\{
  j :
  \frac{ c_j(S)}{|S|}
  >
  \frac{1}{2}+\frac{\lambda}{2|S|}
  \right\},
\end{equation}
{we have}
\begin{equation}
  G_{\lambda}(\alpha,S,T^\star,\Rm_0)
  \geq
  G_{\lambda}(\alpha,S,T,\Rm_0)
  \qquad \text{for any } T.
\end{equation}
{Symmetrically,} for fixed $T${,} the {corresponding update $S^\star$ satisfies}
\begin{equation}
  G_{\lambda}(\alpha,S^\star,T,\Rm_0)
  \geq
  G_{\lambda}(\alpha,S,T,\Rm_0)
  \qquad \text{for any } S.
\end{equation}
{Thus, starting from an initial seed, alternating these two updates
yields an alternating-optimization procedure whose} gain is {non-decreasing at
each step.}

\paragraph{Minimum-density floor $d_{\min}$ (binary matrices).}
For {a current residual $\Rm_0\in\{0,1\}^{n\times n}$ and sign $\alpha=+1$}, entries $(i,j)\in S\times T$ with $\Rm_{0}(i,j)=0$ are
\emph{holes}: subtracting $\mathbf{1}_S\mathbf{1}_T^\top$ from $\Rm_0$ sets
them to $-1$, {adding nonzero residual entries at those positions.}
The hole count equals exactly $|S||T|(1-d_{\Rm_0}(S,T))$, so controlling
$d_{\Rm_0}(S,T)$ directly limits holes.

We impose a per-node density floor: include $j$ only if
{$ c_j(S,\alpha)/|S|>d_{\min}$}.
Combined with~\eqref{eq:halfstep_deriv}, this gives
\begin{equation}
  \label{eq:ao_update_appx}
  2\alpha\sum_{i\in S}\Rm_{0}(i,j)
  \;>\;
  \max\!\bigl(|S|+\lambda,\;\;2d_{\min}|S|\bigr),
\end{equation}

Setting $d_{\min}=0$ recovers the pure half-step.
{Averaging gives $\alpha d_{\Rm_0}(S,T)>d_{\min}$ for a nonempty} AO {output. For a binary current residual and positive sign, this bounds} holes by $|S||T|(1-d_{\min})$.

\paragraph{Neighbourhood initialisation.}
{Seeds are ranked separately by positive and negative residual. We take up to $M$ seeds per sign and orientation (row or column), giving at most $4M$ starts per extraction. Algorithm~\ref{alg:sbd} shows the row-seed case.}

For {sign $\alpha$, we expand the neighbourhood along entries with $\alpha\Rm_0(i,j)>0$. A row} seed {first reaches columns, then rows, then columns again; a column seed uses the transposed initialisation. We search both seed types. AO updates rows first for a} row {seed, and columns first for a column seed. Before AO, we trim} the {candidate sets using alternating} row {and column updates. If either} set is {empty,} the {initial window is kept. Trimming reduces search cost but can exclude positive-gain blocks. AO uses residual weights within} the {retained sets. Acceptance checks} the {full gain.}

\subsection{Positive-gain certificates}
\label{app:positive_gain}

The following conditions identify seeds that can yield positive-gain. Let $T_0\subseteq V$ be the current nonempty column set and let $d=|T_0|$. For each row $i$, define
\[
c_i(T_0,\alpha)
=
\alpha\sum_{j\in T_0}R(i,j).
\]
The row update keeps $i$ whenever
\[
2c_i(T_0,\alpha)
>
h_d,
\qquad
h_d=\max\{d+\lambda,\,2d_{\min}d\}.
\]

\begin{proposition}[Shared-neighborhood certificate]
\label{prop:shared_neighborhood_guarantee}
Assume that there is a nonempty set $P\subseteq U$ such that
\[
c_i(T_0,\alpha)\ge c_{\min}
\qquad
\text{for all }i\in P,
\]
with
\[
2c_{\min}>h_d
\]
and
\[
|P|(2c_{\min}-d-\lambda)>\lambda d.
\]
Then the next row update has positive gain. Later updates {preserve positive gain if they maximise} the gain with the other support fixed. {An active density floor need not preserve this property.}
\end{proposition}

\begin{proof}
Every row in $P$ passes the update threshold. At fixed $T_0$, the gain can be written as
\[
G_\lambda(\alpha,S,T_0,R)
=
\sum_{i\in S}
\left(
2c_i(T_0,\alpha)-d-\lambda
\right)
-\lambda d.
\]
Let $S_1$ be the row support returned by the update. Since every selected row has positive contribution,
\[
\begin{aligned}
G_\lambda(\alpha,S_1,T_0,R)
&\ge
\sum_{i\in P}
\left(
2c_i(T_0,\alpha)-d-\lambda
\right)
-\lambda d
\\
&\ge
|P|(2c_{\min}-d-\lambda)-\lambda d
>0.
\end{aligned}
\]
Each later exact coordinate update maximizes the gain with the other support fixed. Hence the gain cannot decrease.
\end{proof}

For the initialization $T_0=\mathcal N(u)$ and $\alpha=+1$,
\[
c_i(T_0,+1)
=
\sum_{j\in\mathcal N(u)}R(i,j).
\]
For a binary residual, this is the number of neighbors shared by $i$ and the seed $u$.

A second certificate uses the clustering coefficient.

\begin{proposition}[Clustering certificate]
\label{prop:clustering_gain}
\label{app:clustering_gain}
{Let $R$ be the adjacency matrix of a simple undirected binary graph.} Let $u$ have degree $d_u>1$ and clustering coefficient $C_u$. Then the block
\[
S=T=\mathcal N(u)
\]
has positive gain for $\alpha=+1$ if and only if
\[
C_u>
\frac{d_u+2\lambda}{2(d_u-1)}.
\]
\end{proposition}

\begin{proof}
Let $e_u$ be the number of edges between the neighbors of $u$. Then
\[
2e_u=C_ud_u(d_u-1).
\]
For $S=T=\mathcal N(u)$,
\[
G_\lambda(+1,S,T,R)
=
4e_u-d_u^2-2\lambda d_u.
\]
Thus $G_\lambda>0$ if and only if
\[
C_u>
\frac{d_u+2\lambda}{2(d_u-1)}.
\]
\end{proof}

The two certificates have different roles. The shared-neighborhood condition applies to the AO update. The clustering condition shows that the neighborhood block has positive gain, but it does not guarantee that AO finds this block.

\subsection{Exact-recovery guarantees}
\label{app:recovery_guarantees}

{Fix $R$, a sign $\alpha$, and scoped domains $U,V$.} Let $(S^\star,T^\star)$ be a {positive-gain} fixed point of the {actual updates, including the density floor and tie rule.} Let $T_0\subseteq V$ be the nonempty {initial} column set {and define
\[
h_0=\max\{|T_0|+\lambda,\,2d_{\min}|T_0|\}.
\]}
\begin{proposition}[Recovery of a fixed point]\label{prop:ao_recovery}
{Assume
\[
2c_i(T_0,\alpha)
\begin{cases}
>h_0,&i\in S^\star,\\
\le h_0,&i\in U\setminus S^\star.
\end{cases}
\]
One row update followed by one column update recovers $(S^\star,T^\star)$.}
\end{proposition}
\begin{proof}
{The separation condition makes} the first row update {select exactly} $S^\star${. Since} the {target} is a fixed point, the next column update returns {$T^\star$.}
\end{proof}
{If the target also maximises gain over the scoped domain, the recovered block is optimal there. This does not imply optimality outside that domain. Global optimality alone does not ensure} a fixed point of {updates with an active density floor or a different tie rule.}
For a binary {residual} and {$T_0=\mathcal N(u)$, $c_i(T_0,+1)$ counts shared neighbours. Recovery} requires the {threshold} to separate {target rows from all other candidates after scoping.}

\subsection{{Bounds from empty updates}}
{If a column update from nonempty $S$ returns no column in $V$, then
\[
\alpha d_R(S,\{j\})\le\max\left\{\frac12+\frac{\lambda}{2|S|},d_{\min}\right\}
\quad\text{for every }j\in V.
\]
This follows by reversing} the {update inequality and dividing by $2|S|$. An empty row update gives} the {symmetric bound. Checking both signs bounds the absolute density. The result concerns the final remainder only when the update was evaluated on that remainder. It covers only the searched domain.}

\clearpage
\section{{Additional analysis}}
\label{app:additional_analysis}

\subsection{{Local positive-gain structure}}
Table~\ref{tab:theoretical_bounds} reports the {rates at $\lambda=1$. The script evaluates the original binary graph, before scoping, with $T_0=\mathcal N(u)$. Its local test uses $h_d=d+\lambda$, without an active density floor. These rates do not certify the candidates retained by the implemented search.}

\begin{table}[h]
\centering
\caption{{Percentage of nodes satisfying the clustering or shared-neighborhood positive-gain condition at $\lambda=1$. Tests use the original binary graph before scoping and do not apply a density floor. These rates describe available local structure, not the success rate of the implemented search.}}
\small
\begin{tabular}{lrr}
\toprule
Dataset & Clustering (\%) & Shared neighbors (\%) \\
\midrule
Cora & 0.1 & 9.1 \\
PubMed & 0.0 & 12.1 \\
ca-HepPh & 22.5 & 43.7 \\
ca-AstroPh & 30.2 & 57.0 \\
coPapersDBLP & 78.0 & 91.6 \\
coPapersCiteseer & 83.3 & 93.9 \\
COLLAB & 84.3 & 96.5 \\
\bottomrule
\end{tabular}

\label{tab:theoretical_bounds}
\end{table}

The certificate rates are low on Cora and PubMed and high on the coauthorship and copaper graphs. Higher rates generally come with larger CF values. These rates describe local {positive-gain} structure. They are not bounds on the final CF.

\subsection{Impact of the Regularization Parameter $\lambda$}
\label{app:lambda}

We first derive the gain thresholds for several block shapes. We then measure
how changing $\lambda$ affects the residual and compression.

\paragraph{Threshold interpretation.}
Consider a single nonzero residual entry $\Rm_0^{(t)}(i,j) \neq 0$.
{For} a singleton biclique $S_t = \{i\}$, $T_t = \{j\}$, the overlap-corrected
density is {the entry itself, $\tilde{d}_t(\{i\},\{j\}) = \Rm_0^{(t)}(i,j)$, and
the acceptance condition reads}
\begin{equation*}
    |\Rm_0^{(t)}(i,j)| > \tfrac{1}{2}(1 + 2\lambda),
\end{equation*}
which is {exactly the singleton gain condition used in the proof of
Theorem~\ref{thm:sbd_main}, claim~2. Since residual entries are integers,
every nonzero entry satisfies $|\Rm_0^{(t)}(i,j)| \geq 1$, so} for
$\lambda < \tfrac{1}{2}$ every nonzero residual entry yields positive
singleton {gain. Exact recovery then requires the singleton stopping certificate
of Corollary~\ref{cor:sbd_main}.} For $\lambda \geq \tfrac{1}{2}$, {entries of
magnitude $1$ can no longer be removed by a singleton,} and singleton
corrections are never {accepted on residuals with entries in $\{-1, 0, 1\}$.}

\paragraph{Structure-dependent thresholds.}
{The parameter} $\lambda$ controls which {blocks have positive gain. Acceptance requires}
\begin{equation*}
    |\tilde{d}_t(S_t, T_t)| > \frac{1}{2}
    \left(1 + \frac{\lambda}{|S_t|} + \frac{\lambda}{|T_t|}\right).
\end{equation*}
{The threshold is at least $\tfrac{1}{2}$ and increases with $\lambda$.
When residual entries are bounded by $1$ in magnitude (e.g.\ binary $\Am$
before any overlap), we have $|\tilde{d}_t| \leq 1$, so a block of shape
$|S_t| \times |T_t|$} can be {extracted, even when fully dense, only if}
\begin{equation*}
    \frac{\lambda}{|S_t|} + \frac{\lambda}{|T_t|} < 1 .
\end{equation*}
{We call this bound the \emph{extraction limit} for the block shape. Equation~}\eqref{eq:acceptance_condition} {gives:}
\begin{itemize}[leftmargin=0.5cm]
    \itemsep0em
    \item \textbf{Singleton} ($|S_t| = |T_t| = 1$): threshold is
    {$|\tilde{d}_t| > \tfrac{1}{2}(1 + 2\lambda)$; extraction limit
    $\lambda < \tfrac{1}{2}$}.
    \item \textbf{Star} ($|S_t| = 1$, $|T_t| = k$): threshold is
    {$|\tilde{d}_t| > \tfrac{1}{2}(1 + \lambda + \lambda/k)$; extraction
    limit $\lambda < \tfrac{k}{k+1}$}. For large $k$, the threshold
    approaches {$\tfrac{1}{2}(1+\lambda)$}, {below the singleton threshold,}
    making wide stars easier to accept than singletons.
    \item \textbf{$2\times 2$ biclique} ($|S_t| = |T_t| = 2$): threshold is
    {$|\tilde{d}_t| > \tfrac{1}{2}(1 + \lambda)$; extraction limit
    $\lambda < 1$}.
    \item \textbf{$2\times 3$ biclique} ($|S_t| = 2$, $|T_t| = 3$):
    threshold is {$|\tilde{d}_t| > \tfrac{1}{2}(1 + \tfrac{5\lambda}{6})$;
    extraction limit $\lambda < \tfrac{6}{5}$}.
\end{itemize}

\begin{table}[t]
\centering
\caption{{Extraction limits for fully dense blocks when residual entries have magnitude at most one. Larger blocks allow higher $\lambda$.}}
\renewcommand{\arraystretch}{1.3}
\begin{tabular}{lccc}
\toprule
Structure & $|S_t|$ & $|T_t|$ & {Extraction limit} ($\lambda <$) \\
\midrule
Singleton       & 1 & 1 & $\tfrac{1}{2}$  \\
Star ($k = 5$)  & 1 & 5 & $\tfrac{5}{6}$  \\
$2\times 2$     & 2 & 2 & $1$             \\
$2\times 3$     & 2 & 3 & $\tfrac{6}{5}$  \\
\bottomrule
\end{tabular}
\label{tab:lambda_thresholds}
\end{table}

\begin{figure}[h]
    \centering
    \includegraphics[width=\textwidth]{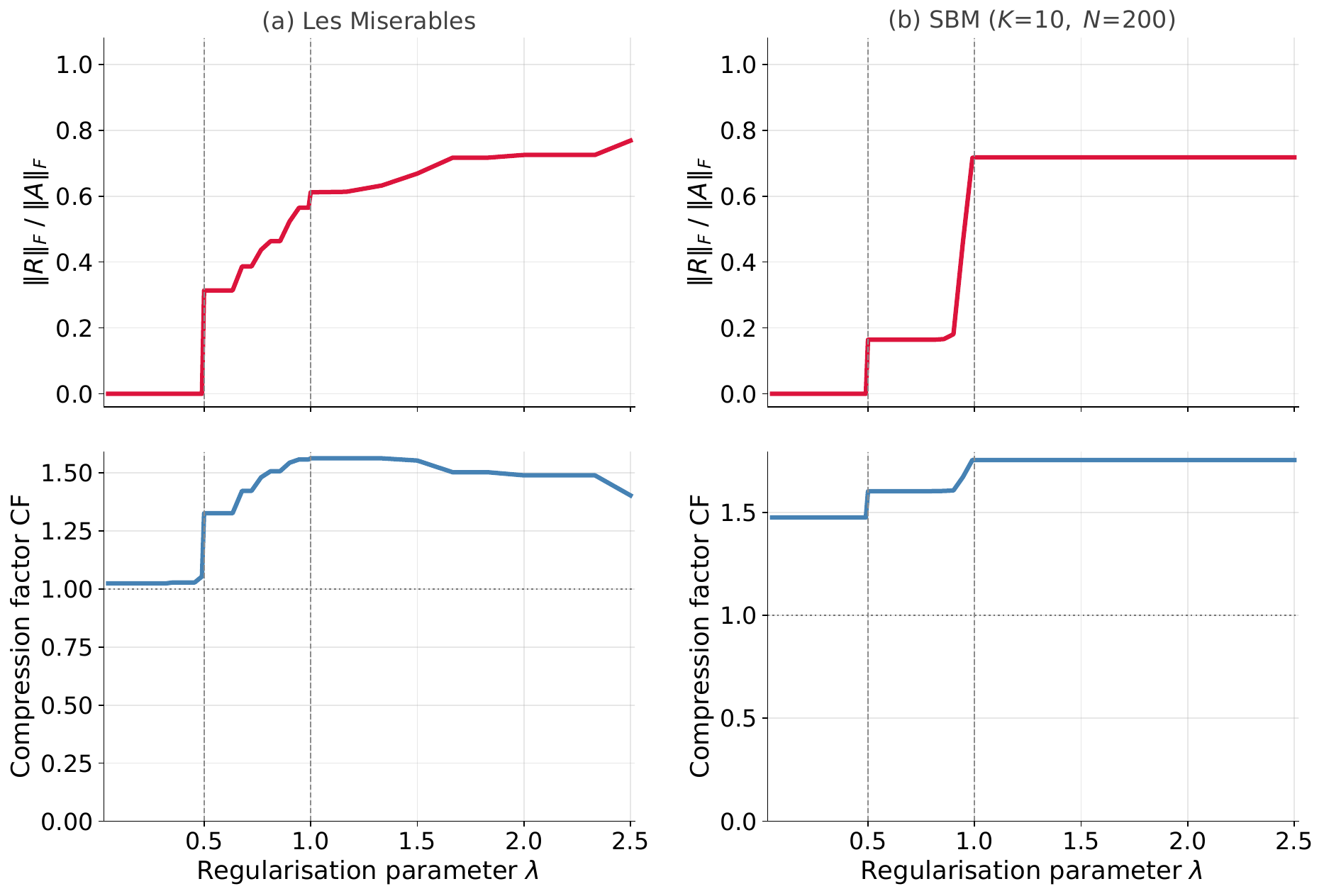}
    \caption{{Normalised residual $\|\Rm_0^{(K)}\|_F/\|\Am\|_F$ and
    compression factor CF as a function of $\lambda$. Dashed vertical lines
    mark $\lambda = \tfrac{1}{2}$ (unit-singleton threshold) and
    $\lambda = 1$ (the range where the storage bound applies).}}
    \label{fig:lambda_residual}
\end{figure}

\begin{figure}[h]
\centering
\includegraphics[width=\linewidth]{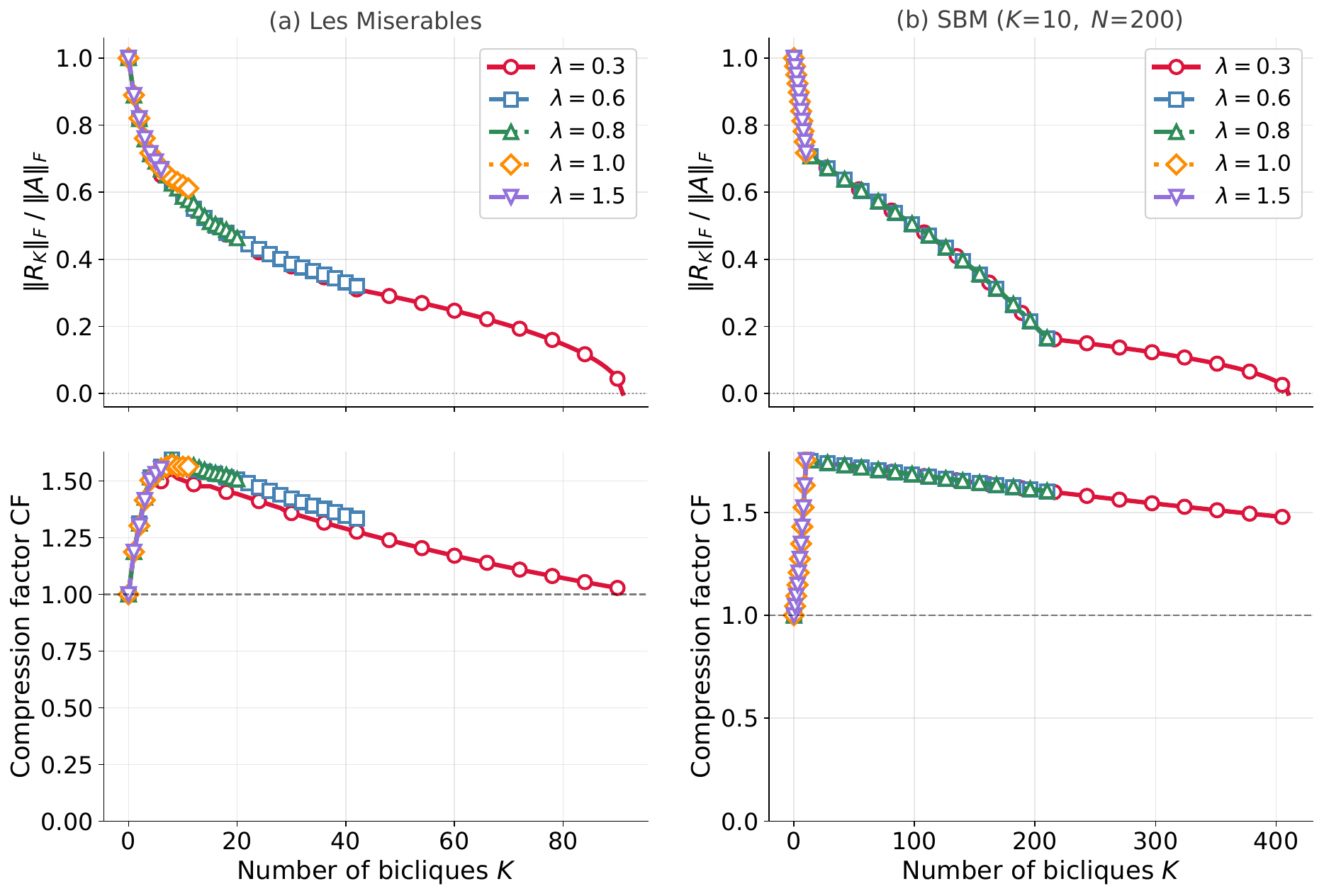}
\caption{{
  Normalised residual $\|\Rm_0^{(K)}\|_F/\|\Am\|_F$ (top row) and
  compression factor CF$_K$ (bottom row) as a function of the number of
  extracted bicliques $K$, for five values of $\lambda$.
  \textbf{(a)} Les~Misérables ($N=77$).
  \textbf{(b)} SBM ($K=10$, $N=200$).
  Dashed horizontal line in bottom row: CF$=1$ (break-even point).
  On the SBM, $\lambda\in\{1.0,1.5\}$ converge after exactly 10 bicliques
  {in this run.}
  {At the plotted $\lambda=0.3$, the residual reaches zero in both graphs.}
  For small $\lambda$, CF declines at large $K$ once the remaining
  extractions are singletons and small stars whose index cost exceeds the
  edges they remove.
}}
\label{fig:app:lambda_kcurves}
\end{figure}

\paragraph{Effect on residual norm.}
{For a fixed residual, increasing} $\lambda$ {raises} the acceptance {threshold.
The search path can change, so} the {final} residual {norm need not increase with $\lambda$.}
\begin{itemize}[leftmargin=0.5cm]
\itemsep0em
\item {\textbf{$\lambda<\tfrac12$}}: {every nonzero integer entry gives positive} singleton {gain. A stopping certificate covering all signed singleton supports therefore gives $\Rm_0^{(K)}=0$ by Corollary~\ref{cor:sbd_main}. A finite seed search or budget limit alone does not give this certificate.}
\item {\textbf{$\lambda\in[\tfrac12,1)$}: unit entries cannot be removed by singleton updates. When residual entries have magnitude at most one, a star with $k$ leaves requires $\lambda<k/(k+1)$. Larger dense blocks may still be accepted. The singleton stopping certificate implies $|\Rm_0^{(K)}(i,j)|\le1$.}
\item {\textbf{$\lambda\ge1$}: when residual entries have magnitude at most one, stars and $2\times2$ blocks} are {rejected, and acceptance requires $1/|S_t|+1/|T_t|<1/\lambda$.} Theorem~\ref{thm:sbd_main} {gives} a lower bound on {CF, not compression optimality or monotone CF.}
\end{itemize}
{Binary input does not ensure bounded residual entries after overlapping updates.
Figures~\ref{fig:lambda_residual} and~\ref{fig:app:lambda_kcurves} show} the {measured trends. Small late blocks can reduce} CF {when their storage} cost exceeds the {residual entries} they remove.

\subsection{{Comparison with a rank-one solver}}
\label{app:ilp_comparison}
We compare the AO heuristics with {a mixed-integer} solver {for each} greedy rank-one {problem. Each sign is solved as a binary quadratic} problem {with a 60-second limit. We save the incumbent, solver status, bound and gap for both signs. A step is certified only when both signs are solved to optimality. Greedy and Fast use the current deterministic seed ordering.}

We run the comparison on Karate Club, Les Mis\'erables, and Dolphins for
\[
\lambda\in\{0,0.5,1,2,4\}.
\]
For each method, we report the number of extracted bicliques $K$, the Compression Factor, and the residual Frobenius norm.

\begin{table}[h]
\centering
\caption{{Rank-one solver and AO heuristics on three small graphs as $\lambda$ varies. We report the number of bicliques $K$, compression factor CF, and residual norm after decomposition. Each signed solver subproblem has a 60-second limit. A dagger marks a decomposition with an uncertified step. Rank-one optima do not imply a global optimum for the final decomposition.}}
\vspace{0.2cm}
\scriptsize
\setlength{\tabcolsep}{2.5pt}
\begin{tabular}{l|ccc|ccc|ccc}
\toprule
& \multicolumn{3}{c|}{\textbf{Rank-one MIQP}} & \multicolumn{3}{c|}{\textbf{SBDGreedy}} & \multicolumn{3}{c}{\textbf{SBDFast}} \\
\cmidrule(lr){2-4} \cmidrule(lr){5-7} \cmidrule(lr){8-10}
$\lambda$ & $K$ & CF & $\|R_0\|_F$ & $K$ & CF & $\|R_0\|_F$ & $K$ & CF & $\|R_0\|_F$ \\
\midrule
\multicolumn{10}{c}{\textbf{Karate}} \\
\midrule
0.0 & 33 & 0.79 & 0.00 & 30 & 0.88 & 0.00 & 33 & 0.63 & 2.00 \\
0.5 & 24 & 0.94 & 2.45 & 24 & 0.94 & 2.45 & 24 & 0.96 & 3.61 \\
1.0 & 9 & 1.17 & 7.48 & 8 & 1.17 & 7.87 & 9 & 1.18 & 7.35 \\
2.0 & 0 & 1.00 & 12.5 & 0 & 1.00 & 12.5 & 0 & 1.00 & 12.5 \\
4.0 & 0 & 1.00 & 12.5 & 0 & 1.00 & 12.5 & 0 & 1.00 & 12.5 \\
\midrule
\multicolumn{10}{c}{\textbf{Les Mis\'erables}} \\
\midrule
0.0 & 88 & 1.03 & 0.00 & 88 & 1.03 & 0.00 & 73 & 0.96 & 5.57 \\
0.5 & 43 & 1.33 & 6.93 & 45 & 1.28 & 6.71 & 50 & 1.23 & 8.12 \\
1.0 & 12 & 1.56 & 13.5 & 12 & 1.54 & 13.7 & 16 & 1.46 & 13.8 \\
2.0 & 4 & 1.50 & 16.2 & 4 & 1.49 & 16.3 & 4 & 1.40 & 17.1 \\
4.0 & 1 & 1.18 & 20.1 & 1 & 1.18 & 20.1 & 1 & 1.15 & 20.4 \\
\midrule
\multicolumn{10}{c}{\textbf{Dolphins}} \\
\midrule
0.0 & 86 & 0.69$^{\dagger}$ & 0.00 & 89 & 0.69 & 0.00 & 79 & 0.62 & 5.29 \\
0.5 & 62 & 0.81$^{\dagger}$ & 5.39 & 61 & 0.80 & 5.57 & 58 & 0.82 & 7.48 \\
1.0 & 14 & 1.09 & 13.6 & 10 & 1.09 & 14.4 & 16 & 1.07 & 13.4 \\
2.0 & 0 & 1.00 & 17.8 & 0 & 1.00 & 17.8 & 0 & 1.00 & 17.8 \\
4.0 & 0 & 1.00 & 17.8 & 0 & 1.00 & 17.8 & 0 & 1.00 & 17.8 \\
\bottomrule
\end{tabular}
\label{tab:lambda_behavior_ilp}
\end{table}

The {methods often give similar} CF and residual {norms, but can return different numbers of blocks. This compares final decompositions; it does} not {establish} the {per-step} gap to {an} exact {oracle or} test Proposition~\ref{prop:ao_recovery}.

\subsection{Why SVD-based Methods are Mismatched to our Objective}
\label{app:slr}

Standard sparse-plus-low-rank methods such as RPCA \citep{RPCA} decompose $\mathbf{A} = \mathbf{L}_0 + \mathbf{R}_0$ where $\mathbf{L}_0$
is a real-valued low-rank matrix and $\mathbf{R}_0$ is sparse.
These are designed for settings where a \emph{dense continuous signal} is corrupted
by sparse noise.
Our objective is different by construction: we seek a decomposition into
\emph{binary integer} rank-one bicliques plus a sparse integer residual.
The two problems differ in what the low-rank
component is allowed to be.

\begin{figure}[h]
\centering
\includegraphics[width=\linewidth]{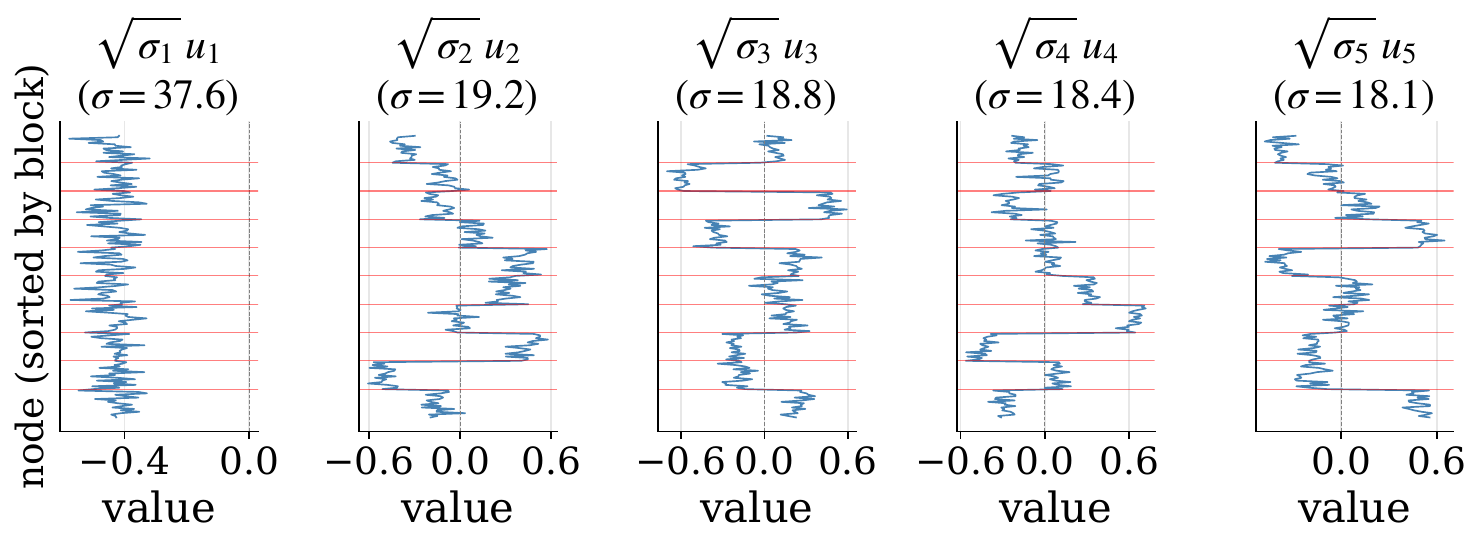}
\caption{{
  Top-$K$ symmetric factors $\sqrt{\sigma_k}\,u_k$ of an SBM adjacency matrix
  ($K=10$, $N=200$, $p_{\mathrm{in}}=1.0$, $p_{\mathrm{out}}=0.1$), nodes sorted
  by ground-truth community (red lines = block boundaries).
  An isolated loopless clique has scaled factor $\sqrt{(B-1)/B}\,\mathbf{1}_{S_k}$; repeated eigenvalues allow mixtures.
  In this example, factors spread across communities,
  producing a real-valued, non-binary vector.
}}
\label{fig:app:slr_eigvecs}
\end{figure}

\paragraph{{SVD factors and community supports.}}
For {an isolated loopless clique} of size $B$, the {leading eigenvalue} is {$B-1$. Its scaled factor is $\sqrt{(B-1)/B}\,\onev_{S_k}$. Equal-sized cliques have repeated eigenvalues, so their eigenvectors can mix. Inter-community edges can also spread the factors across communities, as in Figure~\ref{fig:app:slr_eigvecs}. SVD does not enforce binary factors or} a {sparse remainder.}

\begin{figure}[h]
\centering
\includegraphics[width=.72\linewidth]{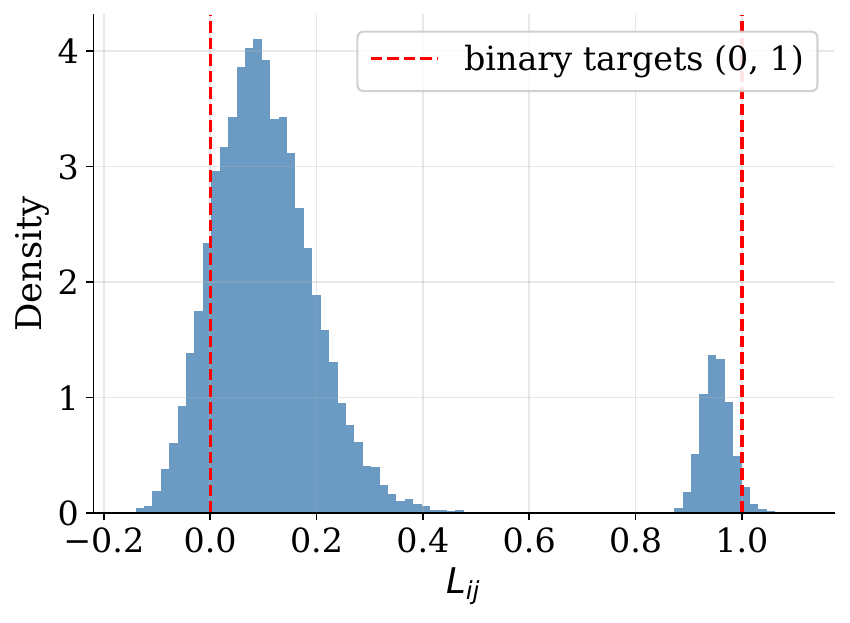}
\caption{{
  Distribution of $\Lm_0^{\mathrm{SparseSVD}}$ entries for 
  SBM ($K=10$, $N=200$, $p_{\mathrm{in}}=1.0$, $p_{\mathrm{out}}=0.1$),
  SparseSVD rank is $10$, $\lambda=0.1$.
}}
\label{fig:app:spca_heatmap}
\end{figure}

\paragraph{{Sparsity on factors does not imply binariness.}}
{Our sparseSVD baseline alternates soft-thresholded row and column updates with rank-one deflation. It} promotes sparsity on the {factors,} which might seem closer to our setting.
However, sparsity does not make the factors binary.
The low-rank component {$\mathbf{L}_0^{\mathrm{SparseSVD}} = \sum_k \sigma_k u_k v_k^\top$}
is a sum of sparse real-valued outer products, entries take arbitrary values in
$\mathbb{R}$, not in $\{0,1\}$.
We verify this even in the most favorable case: SBM with $K=10$, $N=200$,
$p_{\mathrm{in}}=1.0$, $p_{\mathrm{out}}=0.1$, {SparseSVD} rank set to the true $K$,
$\lambda=0.1$.
Even here, \textbf{75.6\%} of entries of {$\mathbf{L}_0^{\mathrm{SparseSVD}}$} exceed 0.05
in {magnitude,} compared to
the original graph's \textbf{18.5\%} nonzeros (Figure~\ref{fig:app:spca_heatmap}).

\paragraph{Scalability.}
On sparse graphs, {each matrix-vector product} costs {$O(E)$. RPCA repeatedly applies singular-value thresholding to dense intermediate matrices,} which {increases memory use and runtime.}
\subsection{{Full kernel comparison}}
\label{app:kernel_comparison_values}
{Table~\ref{tab:rebuttal_kernels_compact} gives the per-width results for Figure~\ref{fig:kernel_comparison}.

Figure~\ref{fig:kernel_comparison} compares SpMM speedups over cuSPARSE. SBD\_base uses the same kernel without bicliques, so the SBD gap measures their effect. The largest gains are on coPapersDBLP, coPapersCiteseer, and COLLAB, which have high CF. On COLLAB, mean SBD speedup is $6.37\times$ at $F=32$, $3.89\times$ at $F=128$, and $3.86\times$ at $F=256$ (Appendix~\ref{app:kernel_comparison_values}). Factor multiplications add overhead on Cora and PubMed. Compression does not guarantee faster SpMM.}
\begin{table}[H]
\centering
\caption{{Mean $\pm$ sample standard deviation of speedups over cuSPARSE across RTX~3090, A40, and L40S. SBD\_base uses no bicliques. All six kernels, including CBM, were measured on each GPU.}}
\scriptsize
\begin{tabular}{lrcccccc}
\toprule
Dataset & $F$ & Sputnik & dgSPARSE & GE-SpMM & SBD\_base & CBM & SBD \\
\midrule
Cora & 32 & $2.72_{\pm 0.37}$ & $1.25_{\pm 0.16}$ & $2.20_{\pm 0.18}$ & $1.70_{\pm 0.28}$ & $0.17_{\pm 0.06}$ & $1.45_{\pm 0.19}$ \\
Cora & 128 & $2.86_{\pm 0.62}$ & $1.41_{\pm 0.12}$ & $2.34_{\pm 0.53}$ & $1.75_{\pm 0.40}$ & $0.19_{\pm 0.08}$ & $1.64_{\pm 0.26}$ \\
Cora & 256 & $2.34_{\pm 0.35}$ & $1.21_{\pm 0.04}$ & $2.12_{\pm 0.47}$ & $1.49_{\pm 0.21}$ & $0.18_{\pm 0.06}$ & $1.49_{\pm 0.20}$ \\
\midrule
PubMed & 32 & $2.68_{\pm 0.40}$ & $1.43_{\pm 0.13}$ & $2.83_{\pm 0.70}$ & $1.83_{\pm 0.30}$ & $0.21_{\pm 0.07}$ & $1.68_{\pm 0.31}$ \\
PubMed & 128 & $1.84_{\pm 0.26}$ & $1.25_{\pm 0.02}$ & $1.80_{\pm 0.25}$ & $1.47_{\pm 0.11}$ & $0.25_{\pm 0.10}$ & $1.38_{\pm 0.05}$ \\
PubMed & 256 & $1.73_{\pm 0.37}$ & $1.17_{\pm 0.10}$ & $1.86_{\pm 0.25}$ & $1.39_{\pm 0.02}$ & $0.34_{\pm 0.12}$ & $1.34_{\pm 0.06}$ \\
\midrule
ca-HepPh & 32 & $2.21_{\pm 0.29}$ & $0.91_{\pm 0.03}$ & $1.72_{\pm 0.36}$ & $1.11_{\pm 0.25}$ & $0.19_{\pm 0.06}$ & $1.39_{\pm 0.28}$ \\
ca-HepPh & 128 & $1.70_{\pm 0.16}$ & $0.91_{\pm 0.01}$ & $1.20_{\pm 0.16}$ & $1.03_{\pm 0.10}$ & $0.23_{\pm 0.05}$ & $1.47_{\pm 0.11}$ \\
ca-HepPh & 256 & $1.60_{\pm 0.09}$ & $0.85_{\pm 0.08}$ & $1.24_{\pm 0.14}$ & $1.05_{\pm 0.09}$ & $0.31_{\pm 0.06}$ & $1.36_{\pm 0.21}$ \\
\midrule
ca-AstroPh & 32 & $2.00_{\pm 0.19}$ & $1.07_{\pm 0.07}$ & $1.77_{\pm 0.28}$ & $1.15_{\pm 0.18}$ & $0.23_{\pm 0.07}$ & $1.66_{\pm 0.26}$ \\
ca-AstroPh & 128 & $1.40_{\pm 0.17}$ & $1.02_{\pm 0.03}$ & $1.24_{\pm 0.23}$ & $1.11_{\pm 0.17}$ & $0.29_{\pm 0.06}$ & $1.37_{\pm 0.16}$ \\
ca-AstroPh & 256 & $1.32_{\pm 0.19}$ & $0.85_{\pm 0.13}$ & $1.28_{\pm 0.25}$ & $0.94_{\pm 0.07}$ & $0.44_{\pm 0.07}$ & $1.32_{\pm 0.37}$ \\
\midrule
coPapersDBLP & 32 & $1.56_{\pm 0.43}$ & $1.70_{\pm 0.18}$ & $2.13_{\pm 0.04}$ & $1.76_{\pm 0.04}$ & $1.89_{\pm 0.32}$ & $5.03_{\pm 0.67}$ \\
coPapersDBLP & 128 & $0.92_{\pm 0.34}$ & $1.07_{\pm 0.04}$ & $1.73_{\pm 0.15}$ & $1.68_{\pm 0.23}$ & $1.42_{\pm 0.30}$ & $3.23_{\pm 0.94}$ \\
coPapersDBLP & 256 & $0.92_{\pm 0.34}$ & $0.95_{\pm 0.06}$ & $1.72_{\pm 0.16}$ & $1.45_{\pm 0.23}$ & $1.50_{\pm 0.31}$ & $3.21_{\pm 0.96}$ \\
\midrule
coPapersCiteseer & 32 & $2.11_{\pm 0.37}$ & $1.79_{\pm 0.25}$ & $2.12_{\pm 0.05}$ & $1.78_{\pm 0.07}$ & $2.18_{\pm 0.33}$ & $5.80_{\pm 0.74}$ \\
coPapersCiteseer & 128 & $1.26_{\pm 0.24}$ & $1.03_{\pm 0.05}$ & $1.98_{\pm 0.03}$ & $2.15_{\pm 0.06}$ & $1.79_{\pm 0.40}$ & $3.92_{\pm 1.19}$ \\
coPapersCiteseer & 256 & $1.27_{\pm 0.24}$ & $0.92_{\pm 0.09}$ & $1.95_{\pm 0.07}$ & $2.09_{\pm 0.08}$ & $1.98_{\pm 0.45}$ & $3.89_{\pm 1.19}$ \\
\midrule
COLLAB & 32 & $2.21_{\pm 0.27}$ & $1.82_{\pm 0.36}$ & $2.09_{\pm 0.06}$ & $1.75_{\pm 0.16}$ & $2.60_{\pm 0.34}$ & $6.37_{\pm 0.69}$ \\
COLLAB & 128 & $1.26_{\pm 0.29}$ & $0.97_{\pm 0.07}$ & $1.91_{\pm 0.05}$ & $2.45_{\pm 0.05}$ & $1.86_{\pm 0.39}$ & $3.89_{\pm 1.11}$ \\
COLLAB & 256 & $1.27_{\pm 0.28}$ & $0.89_{\pm 0.02}$ & $1.91_{\pm 0.03}$ & $2.78_{\pm 0.05}$ & $2.02_{\pm 0.45}$ & $3.86_{\pm 1.07}$ \\
\bottomrule
\end{tabular}

\label{tab:rebuttal_kernels_compact}\label{tab:speedup_sbd}
\end{table}

\subsection{{Feature width and retained blocks}}
\label{app:k_sweep_analysis}
{On COLLAB,} the {fastest sampled prefix uses 30,242 of 53,646 blocks} at $F=32$, {1,722 at} $F=128$, {and 308 at $F=256$. Adding more blocks can slow down SpMM. These are measurements for} the {plotted configuration, not universal thresholds across GPUs.}
At $F=32$, {eight lanes per warp load} the {eight \texttt{float4} feature groups. At $F=128$, all 32 lanes load one group. This may affect} memory {throughput, but the curves alone do not establish bandwidth saturation or exclude cache effects. Profiling is} needed to {test that explanation.}

\clearpage
\section{{Implementation details}}
\label{app:implementation}

\subsection{SBDFast implementation (binary matrices)}
\label{app:fast_tricks}

SBDFast has two search phases. First, it proposes positive blocks using
 a binary mask of the positive residual entries. Then it searches for
positive and negative blocks using the integer residual. Both phases
maintain the exact residual and use it to check each proposal before
acceptance. Negative entries can arise from holes or overlaps in
positive blocks.

\paragraph{Parallel search and sequential acceptance.}
Workers process seed nodes in batches against a fixed residual.
They return potential blocks, which are ordered by their search gains.
We then check each block in this order. We recompute its gain on the
current residual and accept it if the gain is positive and the density
condition holds. After each acceptance, we subtract the block from the
residual before checking the next proposal. Several blocks can therefore
be accepted from one batch, reducing the number of parallel searches.
In comparison, SBDGreedy accepts only the best block from each search.

\paragraph{Exact residual updates.}
The residual uses sparse row and column arrays, with separate lists for
fill-ins. Each accepted block updates all entries in its support, including
zeros (which can become fill-ins) and entries of the opposite sign.
Compaction merges the extra lists into the sparse arrays and removes zeros. It changes the storage in the algorithm, not the residual.

\paragraph{Search settings.}
The seed budget controls search effort independently of the thread count.
The main compression and runtime experiments use one signed sweep; the
rank-one comparison uses up to 20 signed sweeps. The candidate row limit remains 400.

\paragraph{Guarantees.}
Every accepted block has positive gain on the current exact residual.
Theorem~\ref{thm:sbd_main} therefore gives residual decrease and objective
descent for SBDFast. For integer input and $\lambda\geq0$, there are at most
$\|\Am\|_F^2$ accepted blocks. The search remains heuristic: it does not
certify that no positive-gain block remains. The stopping bounds require
nonpositive gain checks on the final residual. Exact recovery requires the additional certificate in Corollary~\ref{cor:sbd_main}.

\subsection{{Experimental protocol and hyperparameters}}
\label{app:hyperparameters}
\subsubsection{{Denoising}}

\paragraph{{Search and evaluation.}}
{The Optuna sampler is TPE. Each graph has ten blocks of 20 nodes and $p_{\mathrm{in}}=1$. The noise levels are $q\in\{0.05,0.15,0.25,0.35\}$.}

\paragraph{{Search spaces.}}
{SBDGreedy searches $\lambda\in[0.5,20]$ log-uniformly, $d_{\min}\in[0.5,1]$ uniformly, and an integer step cap in $\{5,\ldots,100\}$. It uses 200 seeds and eight threads. RPCA, Kanada, and sparse SVD search $\lambda\in[10^{-3},2]$ uniformly. Sparse SVD fixes rank 10. Each search uses 30 trials. Savalle uses one 30-trial search with $\tau\in[10^{-3},20]$ log-uniformly and $\gamma\in[10^{-3},1]$ uniformly.}

\paragraph{{Denoising solver defaults.}}
{RPCA uses 1,000 iterations and tolerance $10^{-7}$. Kanada uses 1,000 iterations and tolerance $10^{-6}$. Sparse SVD uses 100 power iterations per component and tolerance $10^{-5}$. Kanada uses ADMM penalty $\rho=1$. Sparse SVD uses random initial vectors. Savalle uses 1,000 iterations for tuning and up to 5,000 for evaluation, with tolerance $10^{-6}$. A failed convergence check gets one retry at fixed $\rho=16$. 

SBDGreedy uses up to eight scoping rounds and 20 AO rounds per start. These caps, the seed and thread counts are fixed.}

\begin{table}[H]
\centering
\caption{{Selected denoising parameters for Table~\ref{tab:denoising}}.}
\small
\begin{tabular}{lrrrr}
\toprule
{Parameter} & {$q=0.05$} & {$q=0.15$} & {$q=0.25$} & {$q=0.35$} \\
\midrule
{SBD $\lambda$} & {4.55} & {4.55} & {6.81} & {8.15} \\
{SBD $d_{\min}$} & {0.58} & {0.58} & {0.51} & {0.58} \\
{SBD step cap} & {19} & {19} & {98} & {83} \\
{RPCA $\lambda$} & {0.12} & {0.06} & {0.06} & {0.06} \\
{Kanada $\lambda$} & {0.12} & {0.12} & {0.06} & {0.06} \\
{Sparse SVD $\lambda$} & {0.31} & {0.41} & {0.50} & {0.75} \\
{Savalle $\tau$} & {3.81} & {6.57} & {7.55} & {7.57} \\
{Savalle $\gamma$} & {0.21} & {0.07} & {0.08} & {0.09} \\
\bottomrule
\end{tabular}

\label{tab:denoising_hp}
\end{table}

\subsubsection{{Decomposition runtime}}

\paragraph{{Fixed settings.}} {Both methods use $\lambda=0.5$ and eight threads. SBDGreedy uses at most $2n$ extractions, 64 seeds per sign and orientation, and $d_{\min}=0$. SBDFast uses a nominal seed budget of $2n$, 32 local starts, batches of 32 seeds, one signed sweep, and $d_{\min}=0.7$. Kanada uses $\lambda=0.5$. RPCA uses $1/\sqrt n$. Both allow up to 5,000 iterations. Sparse SVD extracts up to $n$ components, with $\lambda=0.1$ and up to 100 iterations per component. Savalle uses the float64 symmetry-based ADMM solver, with $\tau=\gamma=0.1$, up to 5,000 iterations, and adaptive penalty starting at $\rho=1$. If it does not converge, it retries with fixed $\rho=16$ under the same total time limit. All continuous solvers use a tolerance of $10^{-6}$ with their respective stopping criteria. Each run has a 60-second compute limit.}

\subsubsection{{Compression and cache selection}}

\paragraph{{Search and fixed settings.}}
{Each graph uses 30 TPE trials with sampler seed 42. The search samples $\lambda\in[0.5,4]$ log-uniformly and $d_{\min}\in[0.5,1]$ uniformly. Each trial uses SBDFast with a nominal seed budget of 500,000, 32 local starts, batches of 32 seeds, one signed sweep, and eight threads. We select the highest CF (Table~\ref{tab:compression_hp}). CBM fixes $\alpha=0$.}

\begin{table}[H]
\centering
\caption{{Parameters recorded in the selected decomposition caches used for Table~\ref{tab:decomp_quality} and SpMM.}}
\small
\begin{tabular}{lrr}
\toprule
Dataset & $\lambda$ & $d_{\min}$ \\
\midrule
Cora & 1.31 & 0.53 \\
PubMed & 1.31 & 0.53 \\
ca-HepPh & 1.23 & 0.50 \\
ca-AstroPh & 1.19 & 0.63 \\
coPapersDBLP & 1.51 & 0.67 \\
coPapersCiteseer & 1.11 & 0.74 \\
COLLAB & 1.35 & 0.70 \\
\bottomrule
\end{tabular}

\label{tab:compression_hp}
\end{table}

\subsubsection{{Compression accounting}}
\label{app:storage_bytes}

\paragraph{{Entry-count model.}}
{Let $s=\sum_k|S_k|$, $t=\sum_k|T_k|$, and $r=\|\Rm_0\|_0$. CF counts one unit per membership, sign, and residual entry:
\[
\mathrm{CF}=\frac{\|\Am\|_0}{s+t+K+r}.
\]
CF does not depend on integer width and excludes pointer arrays.}

\paragraph{{Kernel storage.}}
{The input CSR uses $B_A=4(N+1)+8\|\Am\|_0$ bytes. SBD stores int32 pointers, membership indices, and residual indices and values. Signs use int8. Factor values are one and are not stored. The row-factor and residual pointers have length $N+1$. The column-factor pointers have length $K+1$. Thus
\[
B_{\mathrm{SBD}}=4(2N+K+3)+4(s+t)+8r+K,
\qquad \mathrm{CR}_{\mathrm{SBD}}=B_A/B_{\mathrm{SBD}}.
\]
These counts exclude features, outputs, temporary buffers, and allocator overhead.}

\paragraph{{CBM storage.}}
{We retain the authors' accounting: $B_{\mathrm{CBM}}=4(N+1)+8d+8e$, where $d$ counts delta entries and $e$ counts forest edges. This includes delta CSR and two forest-edge arrays. It excludes GPU branch offsets. Neither ratio measures total GPU memory.}

\paragraph{{CBM implementation.}}
{Our CBM wrapper uses a custom OpenMP decomposition and the authors' GPU update kernels. It converts the forest to branch offsets, adjusts the branch count, and skips empty update launches. We fix $\alpha=0$.}

\subsubsection{{SpMM and block-prefix measurements}}

{We evaluate $F\in\{32,128,256\}$ with float32 features. The kernel scripts do not sweep decomposition parameters or kernel configurations. CBM uses $\alpha=0$. Both kernel comparisons use ten warm-up launches and 50 timed launches. Features use seed 42. Reference CSR indices use int32. Every timed kernel is checked against sparse matrix multiplication within numerical tolerance. For the block-prefix experiment, we sort bicliques by area. We request 20 logarithmically spaced prefix lengths from one to all blocks and round them to integers. We rebuild the exact residual for each prefix. We vary $K$ and $F$ without repeating the decomposition search. The script times the reference and candidate in pairs, with defaults of 10 warm-up and 50 timed launches.}

\subsubsection{{Regularization sweep}}
{SBDGreedy uses a 500-step cap, $d_{\min}=0$, and eight threads. We use 64 seeds on Les Mis\'erables and 200 on the SBM. The summary evaluates 36 values of $\lambda$ between 0.05 and 2.5. The trajectories use $\lambda\in\{0.3,0.6,0.8,1,1.5\}$.}

\subsection{CUDA Kernel Design}
\label{app:kernel}

This appendix describes the CUDA kernels used for SpMM. SBD\_base uses
our kernel without bicliques, so comparison with SBD measures the effect
of the decomposition.

\paragraph{Warp mapping.}
The kernel assigns one warp to each output row. Lanes traverse the
same CSR neighbour list and process different groups of four features
using \texttt{float4} loads.

\paragraph{Register accumulation.}
Each lane accumulates its feature groups in registers. Column aggregation
uses additions, signed biclique combination uses additions or subtractions,
and residual accumulation uses FMAs. The kernels use no shared-memory
staging. SBD stores the first-pass output in a global buffer of shape
$K\times F$.

\paragraph{Memory accesses.}
Lanes read contiguous feature groups, giving coalesced accesses to
the feature matrix. Actual memory traffic depends on cache behaviour.

\paragraph{Synchronisation.}
Each warp keeps its partial sums in registers and computes its row
independently. The kernels use no \texttt{\_\_syncthreads()} or
programmer-managed shared memory. This avoids block barriers and
shared-memory bank conflicts. Ready warps can overlap memory requests,
helping to hide memory latency. The two passes execute in order on the
same CUDA stream.

\paragraph{Two-pass computation.}
We compute $\sum_k\alpha_k\wv_k\qv_k^\top\xv$ in two passes.
The first pass (\texttt{QTX}) assigns one warp per biclique and computes
$\zv_k=\sum_{j\in\qv_k}\xv[j]$.
The second pass (\texttt{WSZ\_RX}) assigns one warp per output row and computes
\[
\sum_{k:\,i\in\wv_k}\alpha_k\zv_k + \sum_j R_{ij}\xv[j].
\]
The intermediate column sum $\zv_k$ is computed once and reused by every row in
$\wv_k$. This amortises column aggregation across rows sharing the same
target set.

\paragraph{Performance regimes.}
The largest gains occur on graphs with many shared-neighbourhood blocks.
On weakly compressible graphs, the cost of the two-pass computation can
outweigh the benefit of reusing column sums.
Table~\ref{tab:speedup_sbd} gives the results for all graphs and widths.
\subsection{{SpMM baselines}}
\label{app:spmm_baselines}
{Table~\ref{tab:rebuttal_kernels_compact} compares Sputnik~\citep{sgk_sc2020}, dgSPARSE, GE-SpMM~\citep{9355302}, SBD\_base, CBM, and SBD. SBD\_base uses the raw adjacency matrix without bicliques.}

\subsection{Computational Resources}
\label{app:compute}

\paragraph{Hardware platforms.}
The SpMM experiments use three GPU platforms with a common software stack: PyTorch 2.11.0, Python 3.12 and CUDA 13.0.
The SpMM benchmarks are run on each platform independently.

\begin{table}[h]
\centering
\vspace{-8pt}
\caption{Hardware platforms used in this work.}
\small
\begin{tabular}{lcccl}
\toprule
\textbf{GPU} & \textbf{VRAM} & \textbf{L2 Cache}&\textbf{Host CPU}  \\
\midrule
NVIDIA RTX~3090  & 24\,GB  & 6MB &{Intel Xeon Silver 4116}\\
 NVIDIA A40       & 48\,GB & 6MB& {Intel Xeon Silver 4210}\\
 NVIDIA L40S      & 48\,GB& 96MB&AMD EPYC {9124}\\
\bottomrule
\end{tabular}
\label{tab:hardware}
\vspace{-8pt}
\end{table}

\paragraph{Timing protocol.}
{We report median per-launch latency} measured with {CUDA events. Both comparisons use ten warm-ups and} 50 timed {launches. Inputs stay} on the {GPU. SBD allocates output and intermediate buffers on each call; some baselines preallocate output. Timing excludes input} transfer {and decomposition.}

\paragraph{Decomposition.}
The SBDFast CPU decomposition uses {a pool of \textbf{8 worker threads}} on all platforms,
matching the SLURM allocation (\texttt{--cpus-per-task=8}).
Decomposition time is excluded from all SpMM figures; it is reported
separately in the decomposition tables.

\paragraph{{Reproduction settings.}}
{Appendix~\ref{app:hyperparameters} lists search budgets and timing settings. Total reproduction time depends} on {hardware} and {solver stopping behaviour.} All experiments can be rerun in several hours.
\subsection{Datasets}
\label{app:dataset}

We evaluate our method on a diverse collection of benchmark graph datasets commonly used in graph representation learning and graph neural network (GNN) research. {They} cover {citation, co-authorship, co-purchase, co-paper,} and {collaboration graphs.}

\paragraph{Citation Networks.}
The citation network benchmarks include \textbf{Cora}, \textbf{CiteSeer}, and \textbf{PubMed}, which are standard node classification datasets introduced in early graph learning literature \cite{Sen_Namata_2008}. Cora contains scientific publications classified into machine learning topics, CiteSeer consists of scientific articles grouped into research categories, and PubMed contains biomedical publications from the PubMed database. Nodes correspond to documents and edges denote citation relationships.

\paragraph{Co-authorship Networks.}
We additionally evaluate on the co-authorship datasets \textbf{CA-AstroPh} and \textbf{CA-HepPh}~\cite{Leskovec2007}, where nodes represent authors and edges indicate co-authorship relationships. These datasets are derived from collaboration networks in the astrophysics and high-energy physics communities and are widely used for large-scale graph representation benchmarking.

\paragraph{Co-paper Networks.}
The \textbf{coPapersDBLP} and \textbf{coPapersCiteseer} datasets~\cite{Geisberger2008} represent paper similarity and co-occurrence relationships derived from the DBLP and CiteSeer corpora.

\paragraph{Social Interaction Graphs.}
Finally, we evaluate on the \textbf{COLLAB} dataset from the TUDataset collection \cite{morris2020tudatasetcollectionbenchmarkdatasets}, which is a scientific collaboration network commonly used for graph classification tasks. Graphs correspond to ego-networks of researchers and edges denote collaborative relationships. We use the disjoint union of all graphs in the dataset. This gives the single adjacency matrix reported in the tables.

All datasets are publicly available and have been extensively adopted in prior graph representation learning studies.

\end{document}

%% file: macros.tex
\usepackage{bm}

\newfont{\bbb}{msbm10 scaled 700}

\newcommand{\qv}{{\bf q}}

\newcommand{\wv}{{\bf w}}

\newcommand{\xv}{{\bf x}}

\newcommand{\zv}{{\bf z}}

\newcommand{\onev}{{\bf 1}}

\newcommand{\Am}{{\bf A}}

\newcommand{\Lm}{{\bf L}}
\newcommand{\Mm}{{\bf M}}

\newcommand{\Rm}{{\bf R}}

\newcommand{\Um}{{\bf U}}

\newcommand{\Vm}{{\bf V}}
\newcommand{\Xm}{{\bf X}}

